%% file: DiscreteSVGDUAI.tex
\documentclass[twoside]{article} 
\usepackage[accepted]{aistats2020}
\usepackage{subfigure}
\usepackage{appendix}
\usepackage{qiangstyle}

\begin{document}

\twocolumn[
\aistatstitle{Stein Variational Inference for Discrete Distributions}
\aistatsauthor{Jun Han$^1$\And Fan Ding$^2$\And Xianglong Liu$^2$\And Lorenzo Torresani$^1$\And Jian Peng$^3$\And Qiang Liu$^4$}
\aistatsaddress{$^1$Dartmouth College $^2$Beihang University $^3$UIUC $^4$UT Austin}
]

\begin{abstract}
Gradient-based approximate inference methods, such as Stein variational gradient descent (SVGD) \cite{liu2016stein}, provide simple and general purpose inference engines for differentiable continuous distributions. However, existing forms of SVGD cannot be directly applied to discrete  distributions. In this work, we fill this gap by proposing a simple yet general framework that transforms discrete distributions to equivalent piecewise continuous distributions, on which the gradient-free SVGD is applied to perform efficient approximate inference. The empirical results show that our method outperforms traditional algorithms such as Gibbs sampling and discontinuous Hamiltonian Monte Carlo on various challenging benchmarks of discrete graphical models. We demonstrate that our method provides a promising tool for learning ensembles of binarized neural network (BNN), outperforming other widely used ensemble methods on learning binarized AlexNet on CIFAR-10 dataset. In addition, such transform can be straightforwardly employed in gradient-free kernelized Stein discrepancy to perform goodness-of-fit (GOF) test on discrete distributions. Our proposed method outperforms existing GOF test methods for intractable discrete distributions. 
\end{abstract}

\input{tex/introduction_old.tex}
\input{tex/background.tex}
\input{tex/ourmethod.tex}

\input{tex/experiment.tex}

\input{tex/conclusion.tex}


\bibliographystyle{ieee}
\bibliography{iclr2019_conference}

\newpage
\onecolumn
\input{tex/appendix.tex}

\end{document}

%% file: tex/introduction_old.tex
\section{INTRODUCTION}
Discrete probabilistic models provide a powerful framework for capturing complex phenomenons and patterns, especially in conducting logic and symbolic reasoning. However, probabilistic inference of high dimensional discrete distribution is in general NP-hard and requires highly efficient approximate inference tools.  

Traditionally, approximate inference in discrete models is performed by either Gibbs sampling and Metropolis-Hastings algorithms, or deterministic variational approximation, such as belief propagation, mean field approximation and variable elimination methods \cite{wainwright2008graphical, dechter1998bucket}.  
However, both of these two types of algorithms have their own critical weaknesses: Monte Carlo methods provides theoretically consistent sample-based (or particle) approximation, but are  typically slow in practice, 
while deterministic approximation 
are often much faster in speed, 
but does not provide progressively better approximation like Monte Carlo methods offers. 
New methods that integrate the advantages of the two methodologies is a key research challenge; see, for example, \cite{liu2015probabilistic, lou2017dynamic, ahn2016synthesis}. 

Recently,  
Stein variational gradient descent (SVGD, \cite{liu2016stein}) 
provides a combination of deterministic variational inference with sampling, for the case of \emph{continuous distributions}. The idea is to directly optimize a particle-based approximation of the intractable distributions by following a functional gradient descent direction, yielding both practically fast algorithms and theoretical consistency. However, because SVGD only works for continuous distributions, a key open question is if it is possible to exploit it for more efficient inference of discrete distributions.

In this work, we leverage the power of SVGD for inference of discrete distributions. 
Our idea is to transform discrete  distributions to piecewise continuous distributions, on which gradient-free SVGD, a variant of SVGD that leverages 
a differentiable surrogate distribution to sample non-differentialbe continuous distributions, is applied to perform inference. 
To do so, 
we design a simple yet general framework for transforming discrete distributions to equivalent continuous distributions, 
which is specially tailored for our purpose, 
so that we can conveniently construct differentiable surrogates when applying GF-SVGD.

We apply our proposed algorithm to a wide range of discrete distributions, 
such as Ising models and restricted Boltzmann machines. We find that our proposed algorithm significantly outperforms 
traditional inference algorithms for discrete distributions. In particular, our algorithm is shown to be provide a promising tool for ensemble learn of binarized neural network (BNN) in which both weights and activation functions are binarized.  
Learning BNNs have been shown to be a highly challenging problem, because standard backpropagation cannot be applied. We cast learning BNN as a Bayesian inference problem of drawing a set of samples (which forms an ensemble predictor) of the posterior distribution of weights,
and apply our SVGD-based algorithm for efficient inference. 
We show that our method outperforms other widely-used ensemble methods such as bagging and AdaBoost in achieving highest accuracy with the same ensemble size on the binarized AlexNet.   

In addition, we develop a new goodness-of-fit test for intractable discrete distributions based on gradient-free kernelized Stein discrepancy on the transformed continuous distributions using the simple transform constructed before. Our proposed algorithm outperforms discrete KSD (DKSD, \cite{yang2018goodness}) and maximum mean discrepancy (MMD, \cite{gretton2012kernel}) on various benchmarks.
\paragraph{Related work on Sampling} 
The idea of transforming 
the inference of discrete 
distributions to continuous distributions has been widely studied, 
which, however, mostly concentrates on 
leveraging the power of Hamiltionian Monte Carlo (HMC); see, for example,  \cite{afshar2015reflection, nishimura2017discontinuous, pakman2013auxiliary, zhang2012continuous, dinh2017probabilistic}. 
Our framework of transforming discrete distributions to piecewise continuous distribution is similar to \cite{nishimura2017discontinuous}, 
but is more general and tailored for the application of GF-SVGD. 
{\bf Related work on goodness-of-fit test} Our goodness-of-fit testing is developed from KSD \cite{liu2016kernelized, chwialkowski2016kernel}, which works for differentiable continuous distributions. 
Some forms of goodness-of-fit tests on discrete distributions have been recently proposed such as \cite{martin2017exact, daskalakis2019testing, valiant2016instance}. But they are often model-specialized and require the availability of the normalization constant. \cite{yang2018goodness, gretton2012kernel} is related to ours and will be empirically compared. 

{\bf Outline} Our paper is organized as follows. Section 2 introduces GF-SVGD and GF-KSD. Section 3 proposes our main algorithms for sampling and goodness-of-fit testing on discrete distributions. Section 4 provides empirical experiments. We conclude the paper in Section 5.

%% file: tex/background.tex
\section{STEIN VARIATIONAL GRADIENT DESCENT}
We first introduce SVGD \cite{liu2016stein},  
which provides deterministic sampling but requires the gradient of the target distribution. 
We then introduce gradient-free SVGD and gradient-free KSD \cite{han2018stein}, which can be applied to the target distribution with unavailable or intractable gradient. 

Let $p(\vx)$ be a differentiable density function supported on $\R^d$. The goal of SVGD is to find a set of samples 
$\{ \vx_i\}_{i=1}^n$ (called ''particles'') to approximate $p$ in the sense that 
$$
\lim_{n\to\infty}\frac{1}{n}\sum_{i=1}^n f(\vx_i) =\E_p [f(\vx)],  
$$
for general test functions $f$.  SVGD achieves this by starting 
with a set of particles $\{  \vy_i\}_{i=1}^n$ drawn from any initial distribution, and iteratively updates the particles by  
\begin{align}\label{equ:xxii}
\vy_i  \gets \vy_i +  \epsilon \ff^*(\vy_i),  ~~~~ \forall i = 1, \ldots, n,  
\end{align}
where $\epsilon$ is a step size, and 
$\ff\colon \RR^d \to \RR^d$ is a velocity field  chosen to drive the  particle distribution closer to the target.
Assume the distribution of the particles at the current iteration is $q$, 
and $q_{[\epsilon\ff]}$ is the distribution of the updated particles $\vy^\prime = \vy + \epsilon \ff(\vy)$. 
The optimal choice of $\ff$ can be framed as the following optimization problem:  
\begin{align}\label{equ:ff00}
 \ff^* & \! = \!  \argmax_{\ff \in \F}\!\!  \bigg\{\!\!\!  -   \frac{d}{d\epsilon} \KL(q_{[\epsilon\ff]} ~|| ~ p) \big |_{\epsilon = 0}=\E_{\vy\sim q}[\steinpxtransp  \ff(\vy)] \bigg\},  \notag \\ 
&\!\!\!\!\!\!\!\text{with}~~~ \steinpxtransp \ff(\vy)  = \nabla_{\vy} \log p(\vy) ^\top \ff (\vy)+ \nabla_{\vy}^\top\ff(\vy), 
\end{align}
where $\F$ is a set of candidate velocity fields, $\ff$ is chosen in $\F$ to maximize the decreasing rate on the KL divergence between the particle distribution and the target, and $\steinpx$ is a linear operator 
called \emph{Stein operator} and is formally viewed as a column vector similar to the gradient operator $\nabla_{\vx}$. 

In SVGD, $\F$ is chosen to be the unit ball of a vector-valued reproducing kernel Hilbert space (RKHS) $\H = \H_0 \times \cdots \times \H_0$,
where  $\H_0$ is an RKHS formed by scalar-valued functions associated with a positive definite kernel $k(\vy,\vy')$, that is, 
$\F = \{\ff \in \H \colon ||\ff||_{\H}\leq  1 \}.$
This choice of $\F$ makes it possible to
consider velocity fields in infinite dimensional function spaces while still obtaining computationally tractable solution. 

\cite{liu2016stein} showed that \eqref{equ:ff00} has a simple closed-form solution: 
\begin{align}\label{svgdoptimal} 
\ff^*(\vy') \propto  \E_{\vy\sim q}[\steinpx k(\vy, \vy')], 
\end{align} 
where $\steinpx$ is applied to variable $\vx$. With the optimal form $\ff^*(\vy'),$  the objective in \eqref{equ:klstein00} equals to 
\begin{align}
\label{solvksd}
\!\!\!\!\D_{\F}(q || p) \overset{def}{=} 
\max_{\ff \in \F} \left\{ \E_{\vy \sim q} \left [\steinpxtransp \ff(\vy)\right] \right\}, 
\end{align}
where $\mathbb{D}_\F(q ~||~ p)$ is the kernelized Stein discrepancy (KSD) defined in \cite{liu2016kernelized,  chwialkowski2016kernel}.

In practice, SVGD iteratively update particles $\{\vy_i\}$ by
$\vy_i \gets \vy_i  + \frac{\epsilon}{n} \Delta \vy_i$, where, 
\begin{align}\label{update11}
\!\!\!\!\! \Delta \vy_i =\sum_{j=1}^n [\nabla \log p(\vy_j)k(\vy_j, \vy_i) +  \nabla_{\vy_j}k(\vy_j,\vy_i)].
\end{align}

\paragraph{Gradient-free SVGD} GF-SVGD \cite{han2018stein} extends SVGD to the setting when the gradient of the target distribution does not exist or is unavailable. The key idea is to replace it with the gradient of the differentiable surrogate $\rho(\vy)$ whose gradient can be calculated easily, 
and leverage it for sampling from $p(\vy)$  
using a mechanism similar to importance sampling. 

The derivation of GF-SVGD is based on the following key observation, 
\begin{equation}
\label{obj:iwksd}
 w(\vy)\steinbxtransp\ff(\vy)  = \steinpxtransp \big(w(\vy)\ff (\vy) \big).
\end{equation}
where $w(\vy)=\rho(\vy)/p(\vy).$ Eq. \eqref{obj:iwksd} indicates that the Stein operation w.r.t. $p$, which requires the gradient of the target $p$,  can be transferred to the Stein operator of a surrogate distribution $\rho$, which does not depends on the gradient of $p$. Based on this observation, 
GF-SVGD modifies \label{equ:klstein00} to optimize the following object,
\begin{align}
\label{gradfreeKLmin}
\!\!\!\! \ff^*& 
\!= \argmax_{\ff \in \F} \{ \E_{q} [\steinpxtransp (w(\vx)\ff(\vy))]\}.
\end{align}

Similar to the derivation in SVGD, the optimization problem \eqref{gradfreeKLmin} can be analytically solved; in practice, GF-SVGD derives a gradient-free update as $\vy_i\leftarrow \vy_i+\frac{\epsilon}{n}\Delta \vy_i,$ where
\begin{align}
\label{update222}
\!\!\!\!\!\! \Delta \vy_i \propto
\!\sum_{j=1}^n 
\!w_j \big[\nabla \log \rho(\vy_j) k( \vy_j, \vy_i) + \nabla_{\vy_j} k(\vy_j, \vy_i) \big], 
\end{align}
which replaces the true gradient $\nabla \log p(\vx)$ with a surrogate gradient $\nabla\log \rho(\vx)$, and then uses an importance weight $w_j:=\rho(\vy^j)/p(\vy^j)$ to correct the bias introduced by the surrogate. \cite{han2018stein} observed that GF-SVGD can be viewed as a special case of SVGD with an ``{importance weighted}'' kernel, 
$\wt{k}(\vy, \vy')= \rho(\vy)/p(\vy) k(\vy, \vy') \rho(\vy')/p(\vy'). $
Therefore, GF-SVGD inherits
the theoretical justifications of SVGD  \cite{liu2017stein}. GF-SVGD is proposed to apply to continuous-valued distributions. 

{\bf Gradient-Free KSD} As shown in \cite{han2018stein}, the optimal decrease rate of the $\KL$ divergence in \eqref{equ:klstein00} is 
\begin{equation}
\label{imp:ksd}
\D^2(q, p) = \E_{\vx, \vx'\sim q}[w(\vx)k_{\rho}(\vx, \vx') w(\vx')],  
\end{equation}
where $\kappa_{\rho}(\bd{x},  \bd{x}')$ is defined as,
\begin{align}
\label{imp:kernel}
\!\! \kappa_{\rho}(\bd{x},  \bd{x}')\!\! & = \!\! \bd{s}_{\rho}(\bd{x})^\top k(\bd{x},\bd{x}')\bd{s}_{\rho}(\bd{x}')
+\bd{s}_{\rho}(\bd{x})^\top \nabla_{\bd{x}'}k(\bd{x},\bd{x}') \\ 
 & \!\! +\bd{s}_{\rho}(\bd{x}')^\top \nabla_{\bd{x}}  k(\bd{x},\bd{x}')\!\! +\!\!\nabla_{\bd{x}}\!\cdot\!(\nabla_{\bd{x}'}k(\bd{x}, \bd{x}')), \notag
\end{align}
where $\bd{s}_{\rho}(\bd{x})$ is score function of the surrogate $\rho(\vx).$ Note that in order to estimate the KSD between $q$ and $p$, we only need samples $\{\vx_i\}$ from $q$, $w(\vx)$ and the gradient of $\rho(\vx).$ Therefore, we obtain a form of \emph{gradient-free KSD}.

The goal of this paper is to develop a tool for goodness-of-fit testing on discrete distribution based on gradient-free KSD and a method for sampling on discrete-valued distributions by exploiting gradient-free SVGD.

%% file: tex/ourmethod.tex
\section{MAIN METHOD}
This section introduces the main idea of this work. We first provides a simple yet powerful way to transform the discrete-valued distributions to the continuous-valued distributions. Then we leverage the gradient-free SVGD to sample from the transformed continuous-valued distributions. Finally, we leverage the constructed transform to perform goodness-of-fit test on discrete distributions.


Assume we are interested in sampling from a given discrete distribution $p_*(\vz)$, defined on a finite discrete set $\mathcal Z=\{\va_1,\ldots, \va_K\}$.  We may assume each $\va_i$ is a $d$-dimensional vector of discrete values.
Our idea is to construct a piecewise continuous distribution $p_c(\vx)$ for $\vx\in  \RR^d$, and a map $\Gamma\colon \RR^d \to \mathcal Z$, 
such that the distribution of $\vz = \Gamma(\vx)$ is $p_*$ when $\vx\sim p_c$. In this way, we can apply GF-SVGD on $p_c$ to get a set of samples $\{\vx_i\}_{i=1}^n$ from $p_c$ and apply transform $\vz_i = \Gamma(\vx_i)$ to get samples $\{\vz_i\}$ from $p_*$. 

\begin{mydef}
A piecewise continuous distribution $p_c$ on $\RR^d$ and map $\Gamma: \RR^d 
\to \mathcal{Z}$ is called to form a 
\textbf{continuous parameterization} of $p_*$, if $\vz = \Gamma(\vx)$ follows $p_* $ when $\vx\sim p_c$. 
\end{mydef}
This definition immediately implies the following  result. 
\begin{pro}
 The continuous distribution $p_c$ and $\Gamma$ form a continuous parameterization of discrete distribution $p_*$ on $\mathcal Z =\{\va_1, \ldots, \va_K\}$, iff 
\begin{align}\label{pstar}
p_*(\va_i) = \int_{\RR^d} 
p_c(\vx) \ind [\va_i = \Gamma(\vx)] d\vx, 
\end{align}
for all $i = 1,\ldots, K$. 
Here $\ind(\cdot)$ is the 0/1 indicator function, $\ind(t) =0$ iff $t = 0$  and $\ind(t)=1$ if otherwise. 
\end{pro}

\begin {algorithm}[ht]
\caption {GF-SVGD on Discrete Distributions} 
\label{alg:alg1}  
\begin {algorithmic}
\STATE {\bf Goal}: 
Approximate a given distribution $p_*(\vz)$ ({input}) on a finite discrete set $\mathcal Z$. \vspace{.5em}  
\STATE 1) Decide a base distribution $p_0(\vx)$ on $\RR^d$ (such as Gaussian distribution), and a map $\Gamma\colon \RR^d \to \mathcal Z$ which partitions $p_0$ evenly. Construct a 
{piecewise continuous distribution 
$p_c$} by \eqref{equ:pc}:
$$
p_c(\vx) \propto p_0(\vx) p_*(\Gamma(\vx)). 
$$
\STATE 2) Construct a {differentiable surrogate} of $p_c(\vx)$, for example, by $\rho(\vx) \propto  p_0(\vx) $ or $\rho(\vx) \propto  p_0(\vx) \tilde p_*(\tilde \Gamma(\vx)),$ where $\tilde p_*$ and $\tilde \Gamma$ are smooth approximations of $p_*$ and $\Gamma$, respectively.  \vspace{.5em}  
\STATE 3) Run gradient-free SVGD on $p_c$ with differentiable surrogate $\rho$: starting from an initial $\{\vx_i\}_{i=1}^n$ and repeat 
$$
\vx_i \! \gets\! \vx_i + \frac{\epsilon}{\sum_i w_i}\!\! \sum_{j=1}^n
w_j(\nabla\rho(\vx_j) k(\vx_j, \vx_i)\! + \!\nabla_{\vx_j} \! k(\vx_j, \vx_i)). 
$$
where $w_j = {\rho(\vx_j)}/{p_c(\vx_j)}$, and $k(\vx,\vx')$ is a positive definite kernel.  \vspace{.5em}  
\STATE 
4) Calculate $\vz_i = \Gamma(\vx_i)$ and {\bf output} sample 
$\{\vz_i \}_{i=1}^n$ for approximating discrete target distribution $p_*(\vz)$. 
\end {algorithmic}
\end {algorithm}

\paragraph{Constructing Continuous Parameterizations}
Given a discrete distribution $p_*$, there are many different continuous parameterizations. 
Because \emph{exact} samples of $p_c$ yield \emph{exact} samples of $p_*$ following the definition, 
we should prefer to choose 
continuous parameterizations 
whose $p_c$ is easy to sample using continuous inference method,  GF-SVGD in particular in our method. 
However, 
it is generally difficult to find a theoretically optimal continuous parameterization, 
because it is difficult to quantitatively the notation of 
difficulty of approximate inference by particular algorithms,
and deriving the mathematically optimal 
continuous parameterization may be computationally demanding and requires analysis in a case by case basis. 

In this work, we introduce a simple yet general framework for constructing continuous parameterizations. 
Our goal is not to search for the best possible continuous parameterization for individual discrete distribution, 
but rather to develop a general-purpose framework that works for a wide range of discrete distributions and can be implemented in an automatic fashion. 
Our method also naturally comes with effective differentiable surrogate distributions with which GF-SVGD can perform efficiently.  

\begin{figure}[htb]
\centering
\begin{tabular}{cc}
\includegraphics[width=0.28\textwidth]{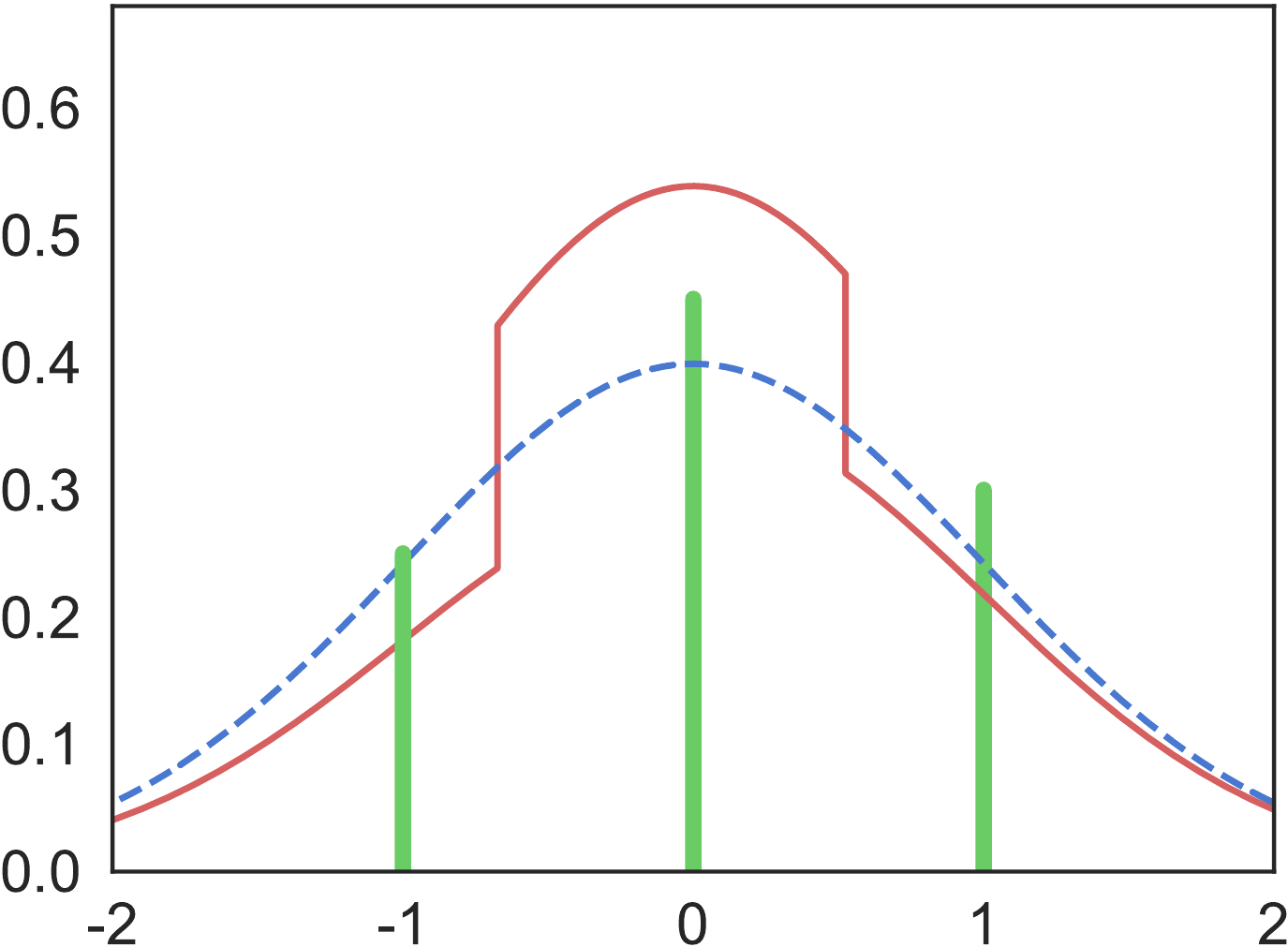} &
\hspace{-.5cm}
\raisebox{3.9em}{\includegraphics[height=0.072\textwidth]{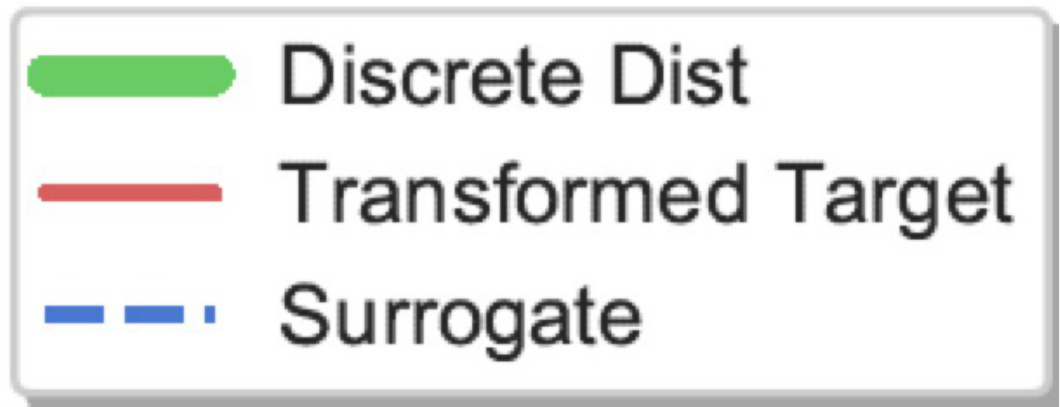}} \\
\end{tabular}
\caption{Illustrating 
the construction of $p_c(\vx)$ (red line) 
of a three-state discrete distribution $p_*$ (green bars). The blue dash line represents the base distribution we use, which is a standard Gaussian distribution.}
\label{fig:definition} 
\end{figure} 

\paragraph{Even Partition} 
Our method starts with choosing a simple base distribution $p_0$, 
which can be the standard Gaussian distribution.  
We then construct a map $\Gamma$ that \emph{evenly partition} $p_0$ into
several regions with equal probabilities. 

\begin{mydef} A map $\Gamma \colon \Z \to \RR^d$ is said to 
\textbf{evenly partition} 
 $p_0$ if we have  
\begin{align} \label{even}
\int_{\RR^d} 
p_0(\vx) \ind[\va_i = \Gamma(\vx)] d\vx = \frac{1}{K}, 
\end{align}
for $i=1,\ldots K$. 
Following \eqref{pstar},   
this is equivalent to saying that 
$p_0$ and $\Gamma$ forms a continuous  relaxation of the uniform distribution $q_*(\va_i) = 1/K$. 
\end{mydef}

For simple $p_0$ such as standard Gaussian distributions, it is straightforward to construct even partitions using the quantiles of $p_0(\vx)$.  
For example,
in the one dimensional case $(d=1)$, we can evenly partition any continuous $p_0(\vx)$, $\vx\in \RR$ by  
\begin{align}\label{equ:gamma1D}
\Gamma(\vx) = \va_i ~~~~~
\text{if ~~ $\vx \in [\eta_{i-1}, ~~ \eta_{i})$}, 
\end{align}
where $\eta_i$ denotes the $i/K$-th quantile of distribution $p_0$. 
In multi-dimensional cases ($d>1$) and when $p_0$ is a product distribution:
\begin{equation}
\label{multiconti:surr}
p_0(\vx) = \prod_{i=1}^d p_{0,i}(x_i).     
\end{equation}

One can easily show that an even partition can be constructed by concatenating one-dimensional even partition: $\Gamma(\vx)=(\Gamma_{1}(x_1),\cdots, \Gamma_{d}(x_d)),$ where $\vx=(x_1,\cdots,x_d)$ and $\Gamma_{i}(\cdot)$ an even partition of $p_{0,i}$.  

A particularly simple case is when $\vz$ is a binary vector, i.e., $\mathcal Z = \{\pm1 \}^d$, in which case $\Gamma(\vx) = \sign(\vx)$ evenly partitions any distribution $p_0$ that is symmetric around the origin. 

\paragraph{Weighting the Partitions} 
Given an even partition of $p_0$, we can conveniently construct 
a continuous parameterization of an arbitrary discrete distribution $p_*$ by 
\emph{weighting each bin of the partition with corresponding probability in $p_*$}, that is, we may  construct $p_c(\vx)$ by 
\begin{align}\label{equ:pc}
p_c(\vx) \propto p_0(\vx) p_*(\Gamma(\vx)),  
\end{align} 
where $p_0(\vx)$ is weighted by $p_*(\Gamma(\vx))$, the probability
of the discrete value $\vz=\Gamma(\vx)$ that $\vx$ maps.

\begin{pro}
Assume $\Gamma$ is an even partition of $p_0(\vx)$, and $p_c(\vx)\propto p_0(\vx) p_*(\Gamma(\vx))$, 
then $(p_c, ~ \Gamma)$ is a continuous parameterization of $p_*$. 
\label{pro4}
\end{pro}

\paragraph{Constructing Differentiable Surrogate} 
Given such a transformation, it is also convenient to construct differentiable surrogate $\rho$ of $p_c$ in \eqref{equ:pc} for GF-SVGD. 
by simply removing $p_*(\Gamma(\vx))$ (so that $\rho=p_0$), 
or approximate it with some smooth approximation, based on properties of $p_*$ and $\Gamma$, that is, 
\begin{align} \label{eq:tildep}
\rho(\vx) =  p_0(\vx) \tilde p_*(\tilde \Gamma(\vx)), 
\end{align}
where $\tilde\Gamma(\cdot)$ denotes 
a smooth approximation of $\Gamma(\vx)$, 
and $\tilde p_*$ is a continuous extension of $p_*(\vx)$ to the continuous domain $\RR^d$. 
See Algorithm~\ref{alg:alg1} for the summary of our main procedure. 

\paragraph{Illustration Using 1D Categorical Distribution}
Consider the 1D categorical distribution $p_*$ shown in Fig.~\ref{fig:definition}, which takes
$\{-1, 0, 1\}$ with probabilities $\{0.25, 0.45, 0.3\}$, respectively.
We use the standard Gaussian base $p_0$ (blue dash), and obtain 
a continuous parameterization $p_c$ using \eqref{equ:pc}, 
in which $p_0(x)$ is weighted by the  probabilities of $p_*$ in each bin.  
Note that $p_c$ is a piecewise continuous distribution. 
In this case, we may naturally choose the base distribution $p_0$ as the differentiable surrogate function to draw samples from $p_c$ when using GF-SVGD. 

\begin{algorithm}[ht]
\caption {Goodness-of-fit testing (GF-KSD)} 
\label{alg:alg2}  
\begin {algorithmic}
\STATE {\bf Input}: Sample $\{\vz_i\}_{i=1}^n\sim q_*$ and its corresponding continuous-valued $\{\vx_i\}_{i=1}^n\sim q_c$, and null distribution $p_c$. Base function $p_0(\vx)$ and bootstrap sample size $m$.
\STATE {\bf Goal}: Test $H_0:q_c=p_c$ vs. $H_1:q_c\neq p_c$.
\STATE -Compute test statistics $\hat{\mathbb{S}}$ by \eqref{emp:ksd}.
\STATE -Compute m bootstrap sample $\hat{\mathbb{S}}^*$ by \eqref{boot:ksd}.
\STATE -Reject $H_0$ with significance level $\alpha$ if the percentage of $\{\hat{\mathbb{S}}^*\}_{i=1}^m$ that satisfies $\hat{\mathbb{S}}^*>\hat{\mathbb{S}}$ is less than $\alpha.$
\end {algorithmic}
\end {algorithm}

\renewcommand{\vy}{\boldsymbol{y}}
\subsection{Goodness-of-fit Test on Discrete Distribution} 
Our approach implies a new method for 
goodness of fit test of discrete distributions, which we now explore. 
Given i.i.d. samples $\{\vz_i\}_{i=1}^n$ from an \emph{unknown} distribution $q_*$, and a candidate discrete distribution $p_*$, 
we are interested in testing $H_0:q_*=p_*$ vs. $H_1:q_*\neq p_*.$ 

Our idea is to transform the testing of discrete distributions $q_* = p_*$ to their continuous parameterizations. 
Let $\Gamma$ be a even partition of a base distribution $p_0$, 
and $p_c$ and $q_c$ are the continuous parameterizations of $p_*$ and $q_*$ following our construction, respectively, that is, 
\begin{align*} 
p_c(\vx) \propto p_0(
\vx) p_*(\Gamma(\vx)), &&
q_c(\vx) \propto p_0(\vx) q_*(\Gamma(\vx)). 
\end{align*}
Obviously, $p_c = q_c$  implies that $p_* = q_*$ (following the definition of continuous parameterization). 
This allows us to transform the problem to a goodness-of-fit test of continuous distributions, which we is achieved by testing if the gradient-free KSD 
\eqref{imp:ksd} equals zero, $H_0:q_c=p_c$ vs. $H_1:q_c\neq p_c.$   

In order to implement our idea, we need to convert the discrete sample $\{\vz_i\}_{i=1}^n$ from $q_*$ to a continuous sample $\{\vx_i\}_{i=1}^n$ from the corresponding (unknown) continuous distribution $q_c$. 
To achieve, note that when $\vx \sim q_c$ and $\vz = \Gamma(\vx)$, 
the posterior distribution $\vx$ of giving $\vz = \va_i$ equals 
$$
q(\vx ~|~ \vz = \va_i) \propto  p_0(\vx) \ind(\Gamma(\vx) = \va_i),
$$
which corresponds to sampling a truncated version of $p_0$ inside the region defined $\{\vx\colon~\Gamma(\vx) = \va_i\}$. 
This can be implemented easily for the simple choices of $p_0$ and $\Gamma$. 
For example,
in the case when $p_0$ is the product distribution in  \eqref{multiconti:surr} 
and $\Gamma$ is the concatenation of the quantile-based partition in  \eqref{equ:gamma1D}, 
we can sample $\vx ~|~ \vz = \va_i$ by sample $\vy$ from $\mathrm{Uniform}([\eta_{i-1}, \eta_i)^d)$ and obtain $\vx$ by $\vx = F^{-1}(\vy)$ where $F^{-1}$ is the inverse CDF of $p_0$. To better understand how to transform the discrete data $\{\vz_i\}_{i=1}^n$ to continuous samples $\{\vx_i\}_{i=1}^n$, please refer to Appendix~\ref{app:gof:transf} for detail.

With the continuous data, the problem is reduced to testing if $\{\vx_i\}_{i=1}^n \sim q_c$  is drawn from $p_c$. 
We achieve this using gradient-free KSD, similar to \cite{liu2016kernelized, chwialkowski2016kernel}. 
In particular, using the surrogate $\rho(\vx)$ in \eqref{eq:tildep}, 
the GF-KSD between the transformed distributions  $q_c$ and $p_c$ is 
\begin{equation}
\label{gof:ksd}
\mathbb{S}(q_c, p_c) =\E_{\bd{x},\vx'\sim q_c}[w(\bd{x})\kappa_{\rho}(\bd{x}, \vx')w(\vx')],
\end{equation}
where $\kappa_{\rho}$ is defined in \eqref{imp:kernel}. 
Under mild conditions~\cite{liu2016kernelized}, it can similarly derived that $\mathbb{S}(q_c, p_c)=0$ iff $q_c=p_c.$ 

With $\{\vx_i\}_{i=1}^n$ from $q_c$, the GF-KSD between $q_c$ and $p_c$ can be estimated by the U-statistics,
\begin{equation}
\label{emp:ksd}
\hat{\mathbb{S}}(q_c, p_c) =\frac{1}{(n-1)n}\sum_{1\le i\neq j\le n} w(\vx_i)\kappa_{\rho}(\bd{x}_i, \vx_j)w(\vx_j).     
\end{equation}
In practice, we can employ the U-statistics $\hat{\mathbb{S}}(q_c, p_c)$ to perform the goodness-of-fit test based on the similar result from \cite{liu2016kernelized, chwialkowski2016kernel}, which replaces their KSD with gradient-free KSD in \eqref{gof:ksd} and follow other procedure. 

{\bf Bootstrap Sample} The asymptotic distribution of $\hat{\mathbb{S}}(q_c, p_c)$ under the hypothesis cannot be evaluated. In order to perform goodness-of-fit test, we draw random multinomial weights $u_1, \cdots, u_n\sim \mathrm{Multi}(n;1/n,\cdots,1/n),$ and calculate
\begin{equation}
\label{boot:ksd}
\hat{\mathbb{S}}^*(q_c, p_c) =\sum_{i\neq j } (u_i\!-\!\frac1n)w(\vx_i) \kappa_{\rho}(\bd{x}_i, \vx_j)w(\vx_j) (u_j-\frac1n).     
\end{equation}
We repeat this process by $m$ times and calculate the critical values of the test by taking the $(1-\alpha)$-th quantile of the bootstrapped statistics $\{\hat{\mathbb{S}}^*(q_c, p_c)\}.$  The whole procedure is summarized in Alg.~\ref{alg:alg2}.

%% file: tex/experiment.tex
\section{EXPERIMENTS}
We apply our algorithm to a number of large scale discrete distributions to  demonstrate its empirical effectiveness. We start with illustrating our algorithm on sampling from a simple one-dimensional categorical distribution. We then apply our algorithm to sample from discrete Markov random field, Bernoulli restricted Boltzman machine. Then we apply our method to learn ensemble models of binarized neural networks (BNN). Finally, we perform experiments on goodness-of-fit test.

\subsection{Statistical Models}
{\bf Ising Model} The Ising model~\cite{ising1924beitrag} is widely used in Markov random field. Consider an (undirected) graph $G=(V, E)$, where each vertex $i\in V$ is associated with a binary spin, which consists of $\vz=(z_1,\cdots,z_d)$. The probability mass function is $p(\vz)=\frac{1}{Z}\sum_{(i,j)\in E} \sigma_{ij}z_iz_j$, $z_i\in\{-1, 1\}$, $\sigma_{ij}$ is edge potential and $Z$ is normalization constant, which is infeasible to calculate when $d$ is high.  

{\bf Bernoulli restricted Boltzmann Machine (RBM)} Bernoulli RBM\cite{hinton2002training} is an undirected graphical model consisting of a bipartite graph between visible variables $z$ and hidden variables $h.$ In a Bernoulli RBM, the joint distribution of visible units $\vz\in \{-1, 1\}^d$ and hidden units $h \in \{-1, 1\}^M$ is given by 
\begin{equation}
\label{def:rbm}
p(\vz, \vh) \propto \exp(-E(\vz, \vh))   
\end{equation}
where $E(\vz, \vh)=-(\vz^\top W \vh+\vz^\top b+\vh^\top c)$, $W\in \mathbb{R}^{d\times M}$ is the weight, $b\in\mathbb{R}^d$ and $c\in\mathbb{R}^M$ are the bias. Marginalizing out the hidden variables $\vh,$ the probability mass function of $\vz$ is given by $p(\vz) \propto \exp(-E(\vz)),$ with free energy $E(\vz)=-\vz^\top b-\sum_k\log(1+\varphi_k),$ where $\varphi_k = \exp(W_{k*}^\top \vz + c_k)$ and $W_{k*}$ is the k-th row of $W.$

\begin{figure*}[htb]
\centering
\begin{tabular}{cccc}
\includegraphics[width=0.24\textwidth]{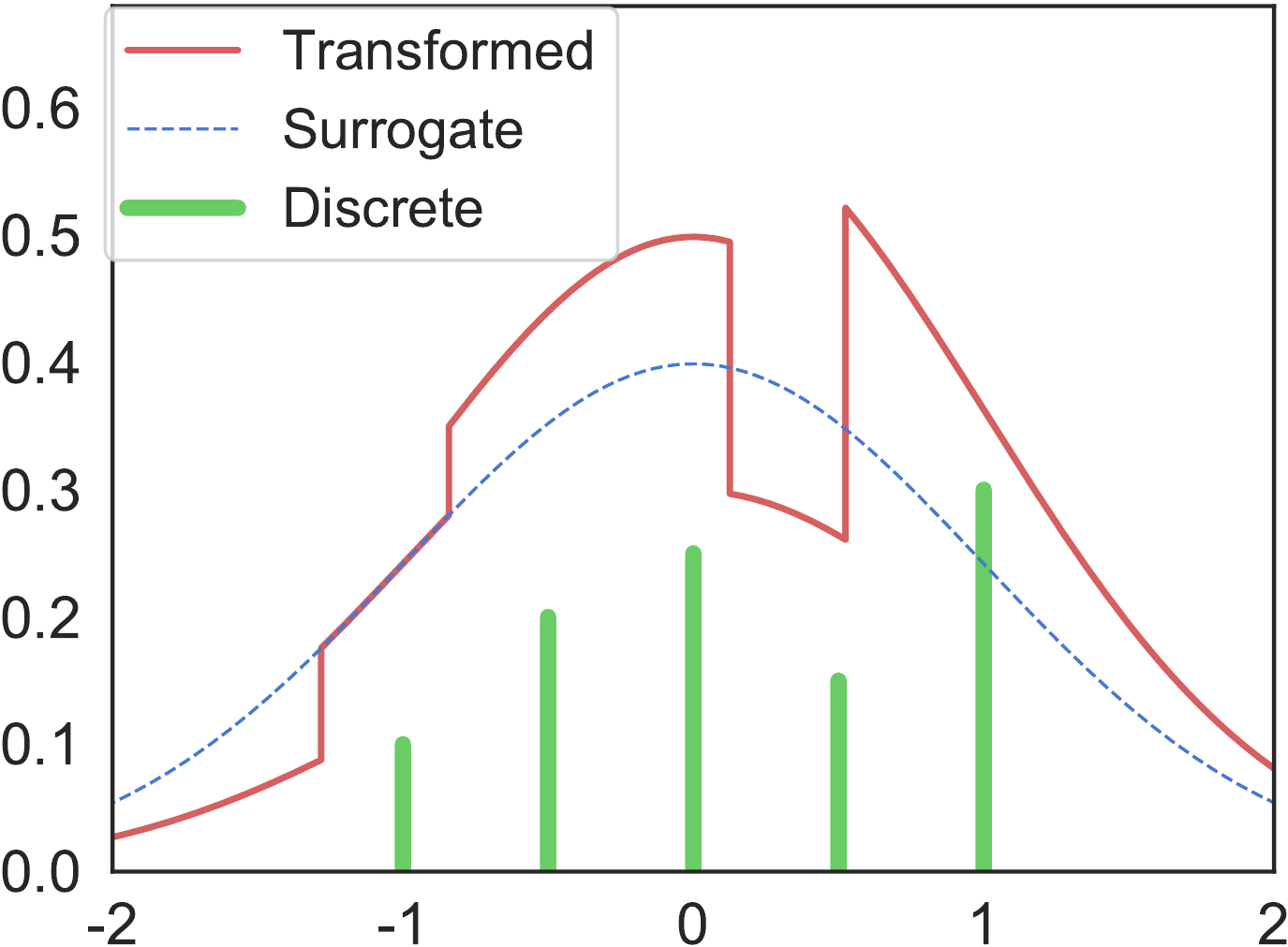} & \hspace{-.5cm}
\includegraphics[width=0.24\textwidth]{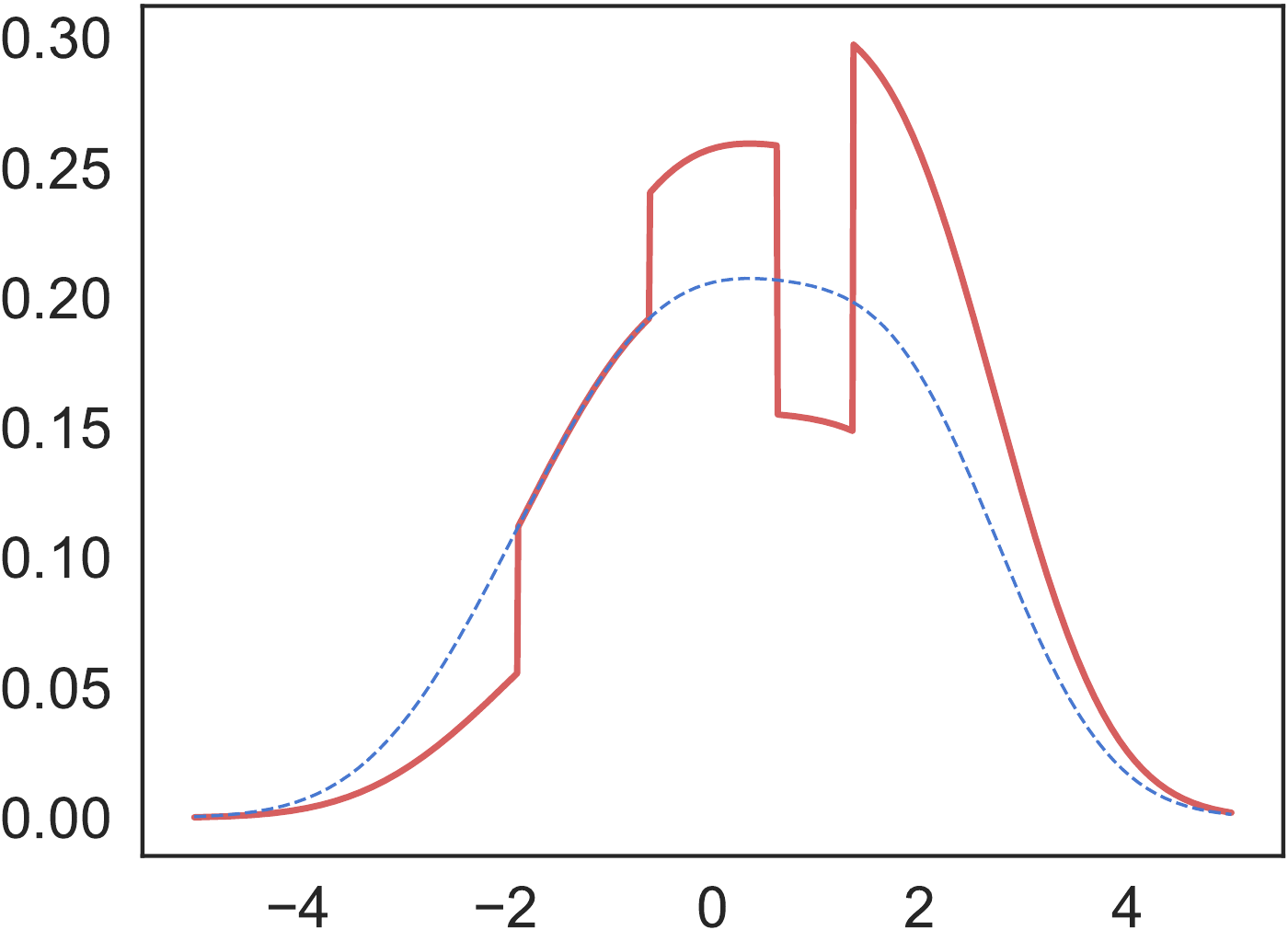} & \hspace{-.5cm}
\includegraphics[width=0.24\textwidth]{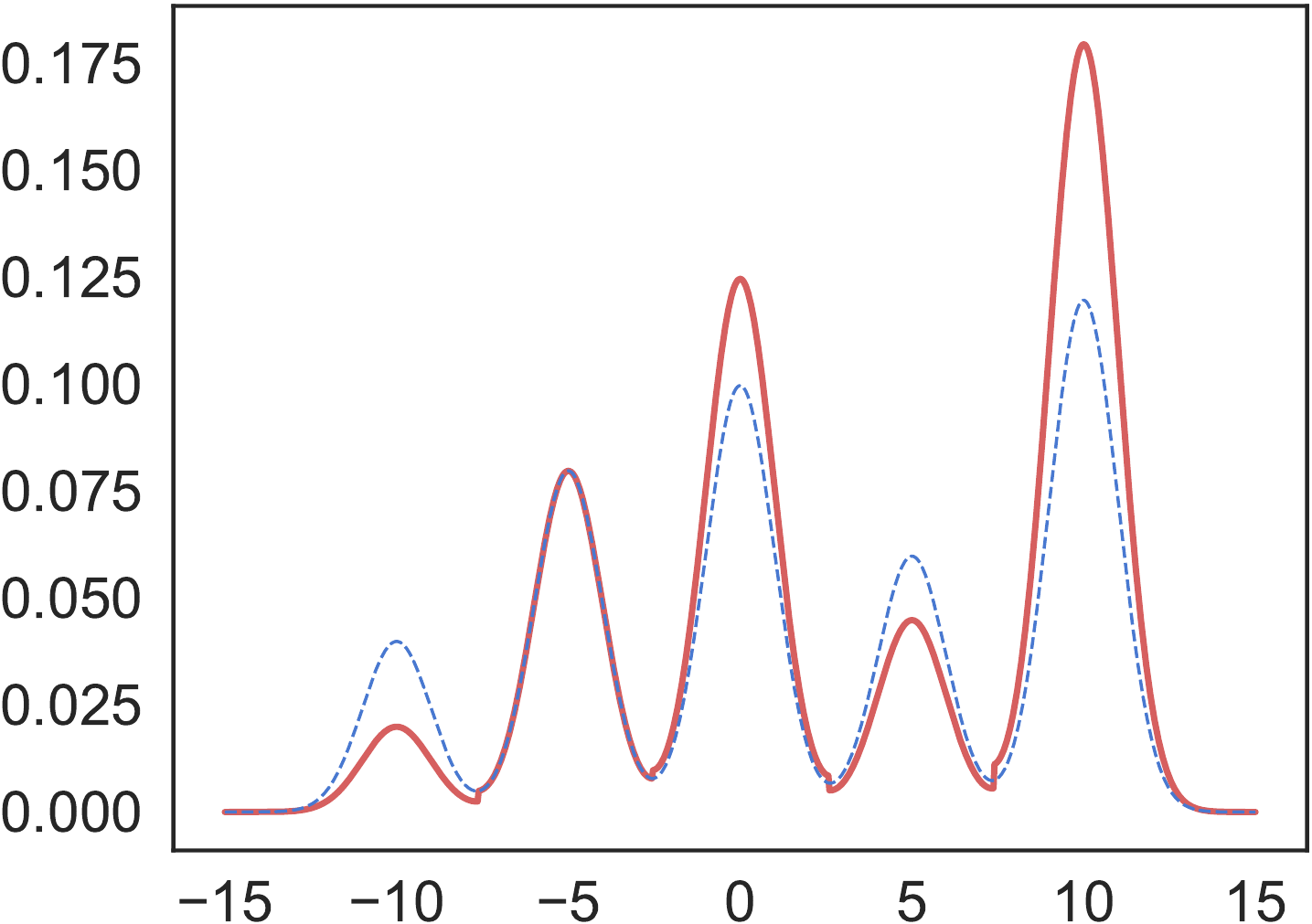} & \hspace{-.5cm}
\includegraphics[width=0.24\textwidth]{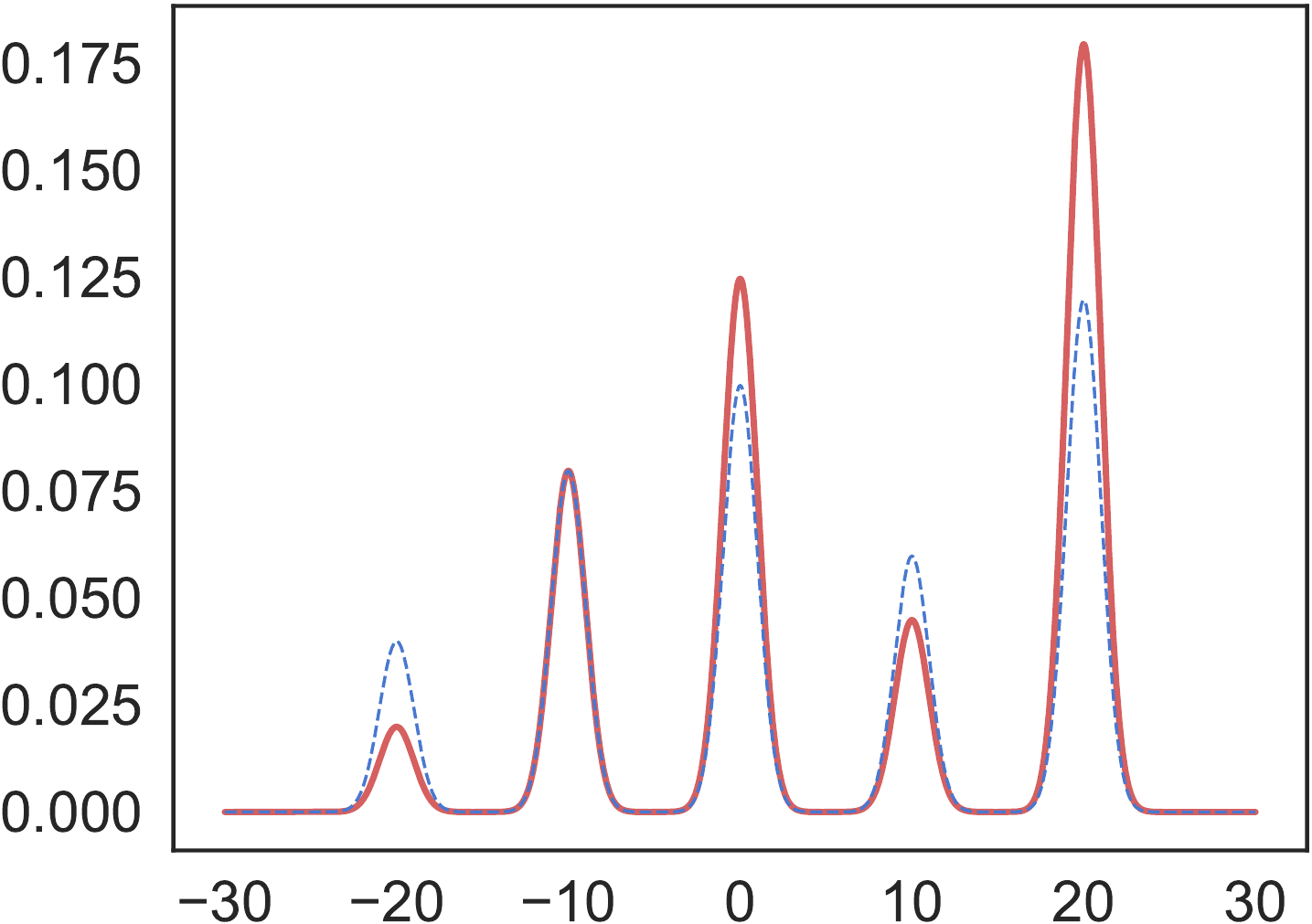} \\
{\small (a), Base $p_0(x)=\mathcal{N}(x; 0, 1)$} & {\small (b), $\bd{\mu}=(-2,-1,0,1,2) $} &
{\small (c), $\bd{\mu}=(-10,-5,0,5,10)$} & {\small (d), $\bd{\mu}=(-20,-10,0,10,20)$} 
\end{tabular}
\caption{Illustrating the construction of $p_c(x)$ (red line) of a five-state discrete distribution $p_*$ (green bars) and the choice of transform. $p_*(z)$ takes values $[-2,-1,0,1,2]$ with probabilities $[p_1, p_2, p_3, p_4, p_5]=[0.1, 0.2, 0.25, 0.15, 0.3]$ respectively. $K=5.$ The dash blue is the surrogate using base $p_0$. Let $p(y)$, $y\in [0, 1)$ be the stepwise density, $p(y\in [\frac{i-1}{K}, \frac{i}{K}))=p_i$, for $i=1,\cdots,K$. In (b, c, d), the base is chosen as $p_0(x)=\sum_{i=1}^5 p_i\mathcal{N}(x; \mu_i, 
1.)$ and $\bd{\mu}=(\mu_1,\mu_2,\mu_3, \mu_4, \mu_5).$ The base $p_0(x)$ in (a) can be seen as $\bd{\mu}=(0., 0., 0., 0., 0.).$
Let $F(x)$ be the c.d.f. of $p_0(x).$ With variable transform $x=F^{-1}(y)$, the transformed target is $p_c(x)=p(F(x)) p_0(x).$ \label{fig:transf}}
\end{figure*}

\subsection{Investigation of the Choice of Transform}
There are many choices of the base function $p_0$ and the transform. We investigate the optimal choice of the transform on categorical distribution in Fig.~\ref{fig:transf}. In Fig.~\ref{fig:transf}(b, c, d), the base is chosen as $p_0(x)=\sum_{i=1}^5 p_i\mathcal{N}(x;\mu_i, 1.0)$ for different $\bd{\mu}.$ The base $p_0(x)$ in Fig.~\ref{fig:transf}(a) can be seen as $\bd{\mu}=(0., 0., 0., 0., 0.).$ We observe that with simple Gaussian base in Fig.~\ref{fig:transf}(a), the transformed target is easier to draw samples, compared with the multi-modal target in Fig.~\ref{fig:transf}(c, d). This suggests that Gaussian base $p_0$ is a simple but powerful choice as its induced transformed target is easy to sample by GF-SVGD.  

\begin{figure}[ht]
\begin{tabular}{ccc}
\includegraphics[width=0.18\textwidth]{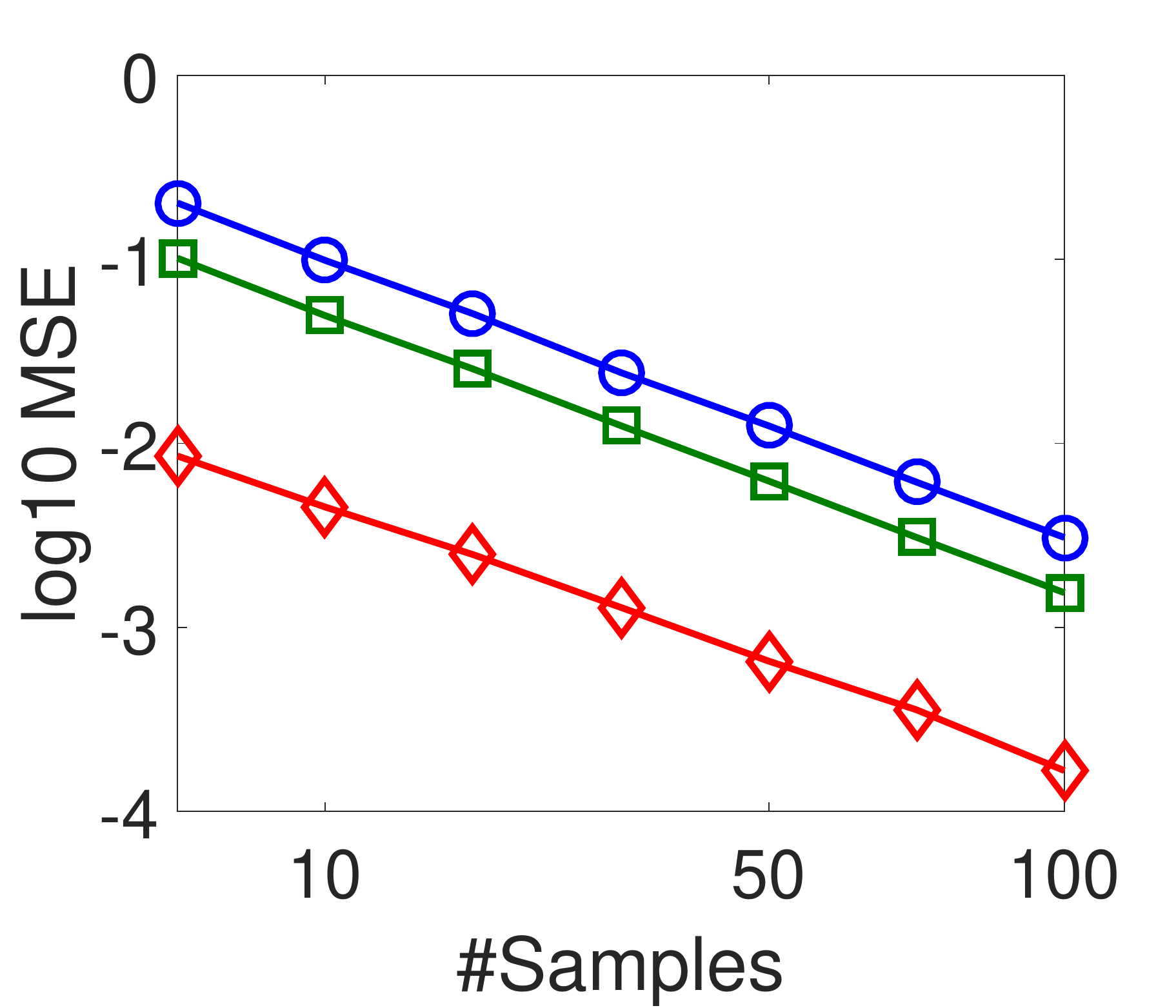} &
\hspace{-0.6cm}
\includegraphics[width=0.18\textwidth]{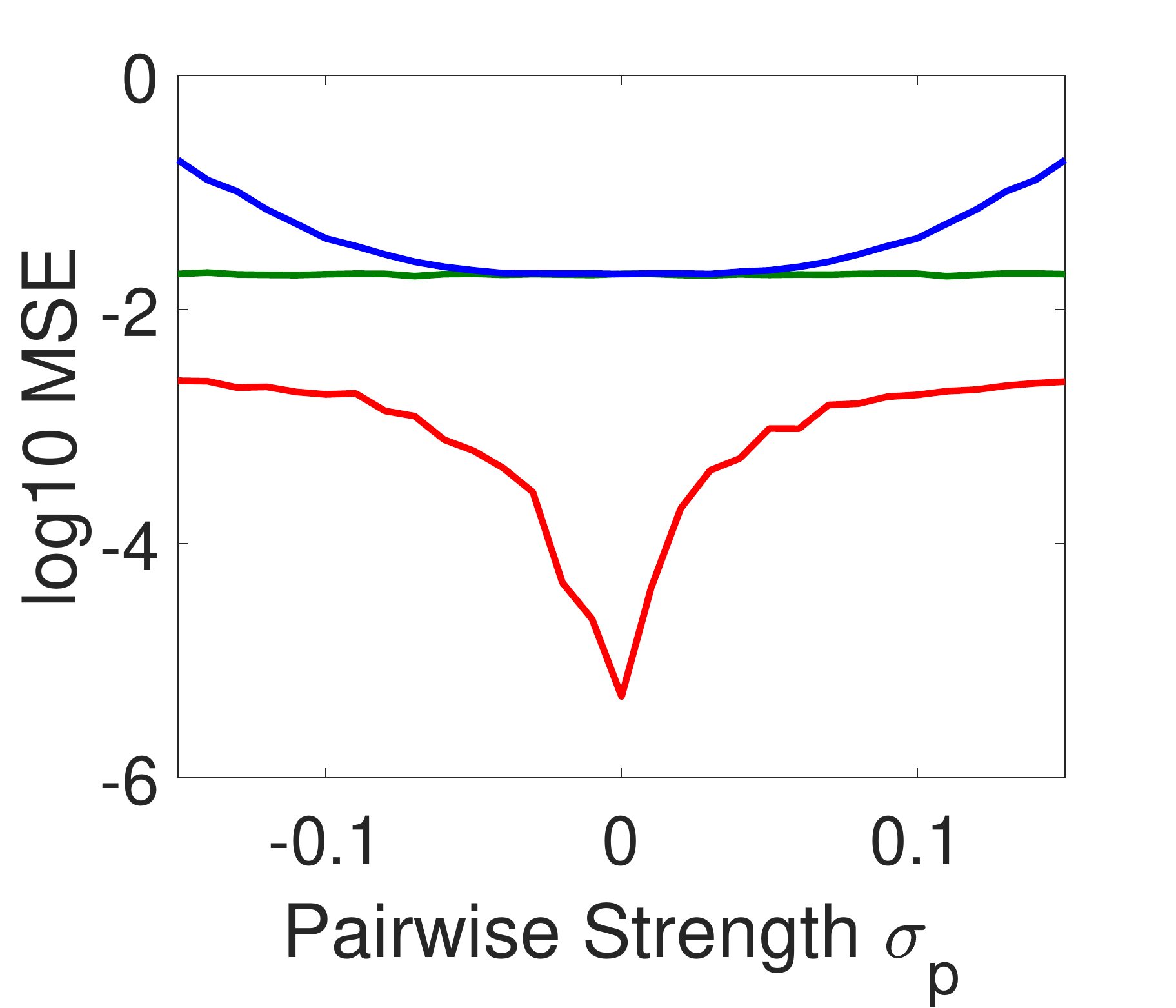} &
\hspace{-0.6cm}
\raisebox{-.5em}{\includegraphics[width=0.1\textwidth]{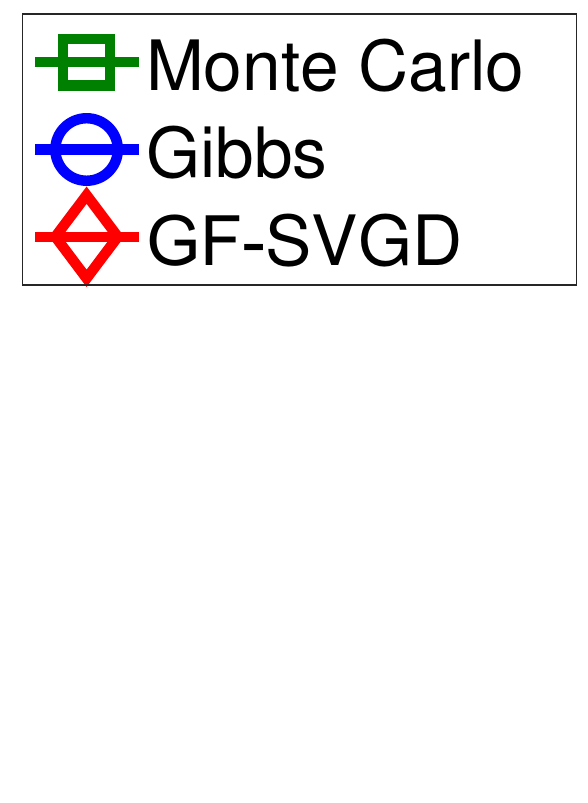}} \\
{\small (a) Fixed $\sigma_p$} &
{\small (b) Fixed $n$} &
{}
\end{tabular}
\caption{Performance of different methods on the Ising model with $10\times10$ grid. We compute the MSE for estimating $\E[\vz]$ in each dimension. Let $\sigma_{ij}=\sigma_{p}.$ In (a), we fix $\sigma_p=0.1$ and vary the sample size $n$. In (b), we fix the sample size $n=20$ and vary $\sigma_p$ from -0.15 to 0.15. In both (a) and (b) we evaluate the $\log~\mathrm{MSE}$ based on 200 trails. DHMC has similar performance as Monte Carlo and is omitted for clear figure.}
\label{fig:ising}
\end{figure}

\subsection{Experiments on Sampling}

{\bf Ising Model} We evaluate the mean square error (MSE) for estimating the mean value $\E_{p_*}[\vz]$ in each dimension. As shown in Section 3, it is easy to map $\vz$ to the piecewise continuous distribution of $\vx$ in each dimension. We take $\Gamma(\vx)=\sign(\vx)$, with the  transformed target $p_c(\vx)\propto p_0(\vx)p_*(\sign(\vx))$. The base function $p_0(\vx)$ is taken to be the standard Gaussian distribution on $\RR^d$.  
We apply GF-SVGD to sample from $p_c(\vx)$ with the surrogate $\rho(\vx)=p_0(\vx)$. The initial particles $\{\vx_i\}$ is sampled from $\mathcal{N}(-2, 1)$ and update $\{\vx_i\}$ by 500 iterations. We obtain $\{\vz_i\}_{i=1}^n$ by $\vz_i=\Gamma(\vx_i)$, which approximates the target model $p_*(\vz)$. We compared our algorithm with both exact Monte Carlo (MC) and Gibbs sampling which is iteratively sampled over each coordinate and use same initialization (in terms of $\vz=\Gamma(\vx)$) and number of iterations as ours. 

Fig.~\ref{fig:ising}(a) shows the log MSE over the log sample size. With fixed $\sigma_s$ and $\sigma_p$, our method has the smallest MSE and the MSE has the convergence rate $\mathcal{O}(1/n)$. The correlation $\sigma_{p}$ indicates the difficulty of inference. As $|\sigma_{p}|$ increases, the difficulty increases. As shown in Fig.~\ref{fig:ising}(b), our method can lead to relatively less MSE in the chosen range of correlation. It is interesting to observe that as $\sigma_p\rightarrow0$, our method significantly outperforms MC and Gibbs sampling.

\paragraph{Bernoulli Restricted Boltzmann Machine}
The base function $p_0(\vx)$ is the product of the p.d.f. of the standard Gaussian distribution over the dimension $d.$ Applying the map $\vz=\Gamma(\vx)=\sign(
\vx)$, the transformed piecewise continuous target is $p_c(\vx)\propto p_0(\vx)p_*(\sign(\vx)).$ Different from previous example, we construct a simple and more powerful surrogate distribution $\rho(\vx)\propto \wt{p}(\sigma(\vy))p_0(\vx)$ where $\wt{p}(\sigma(\vy))$ is differentiable approximation of $p_*$ and $\sigma$ is defined as 
\begin{equation}
\label{binary:approx}
\sigma(\vx)=\frac{2}{1+\exp(-\vx)}-1,    
\end{equation} 
and $\sigma(\vx)$ approximates $\sign(\vx).$ Intuitively, it relaxes $p_c$ to a differentiable surrogate with tight approximation.

\begin{figure}[hbt]
\centering
\begin{tabular}{cc}
\includegraphics[height=0.145\textwidth]{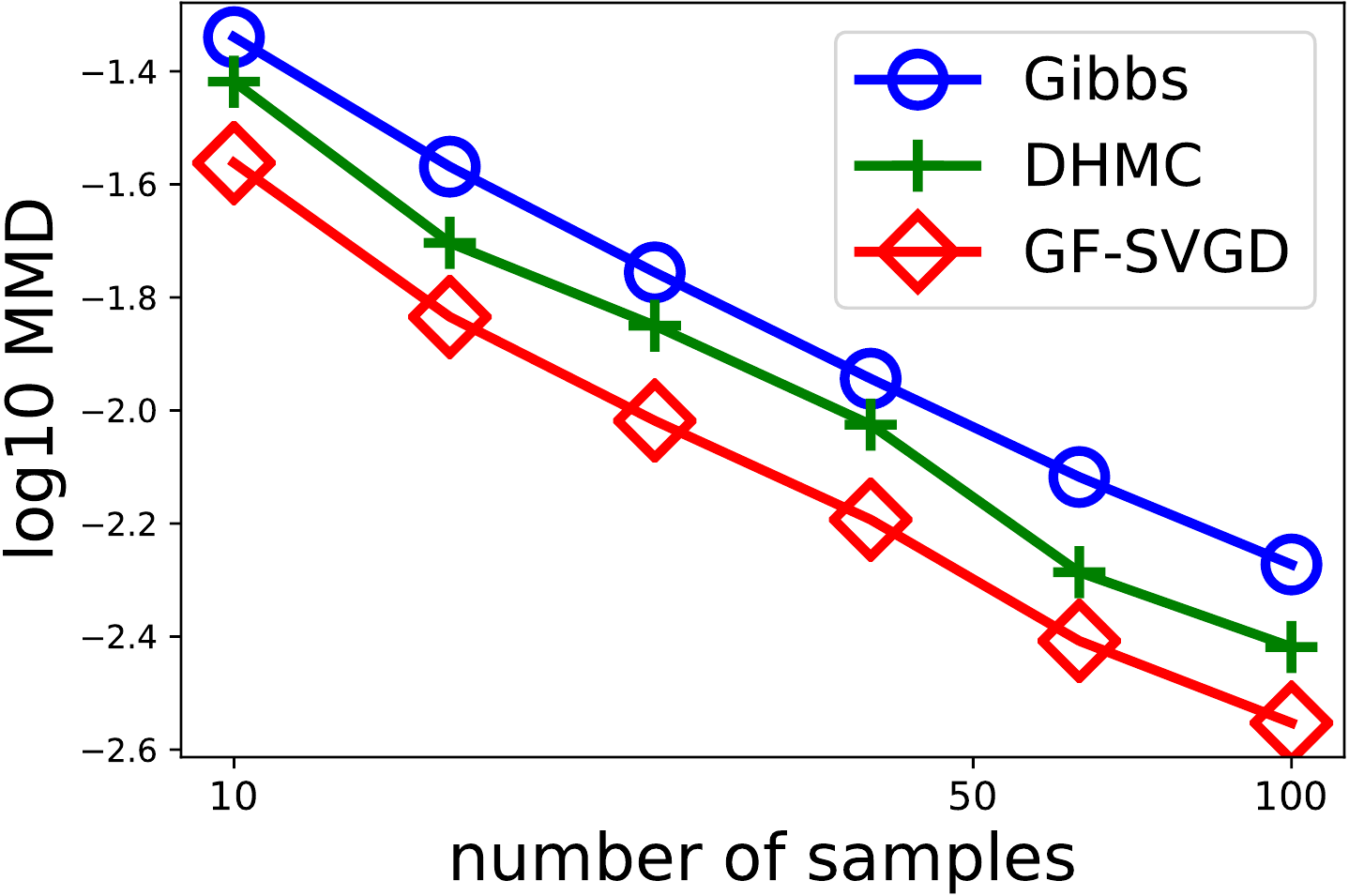} &
\includegraphics[height=0.145\textwidth]{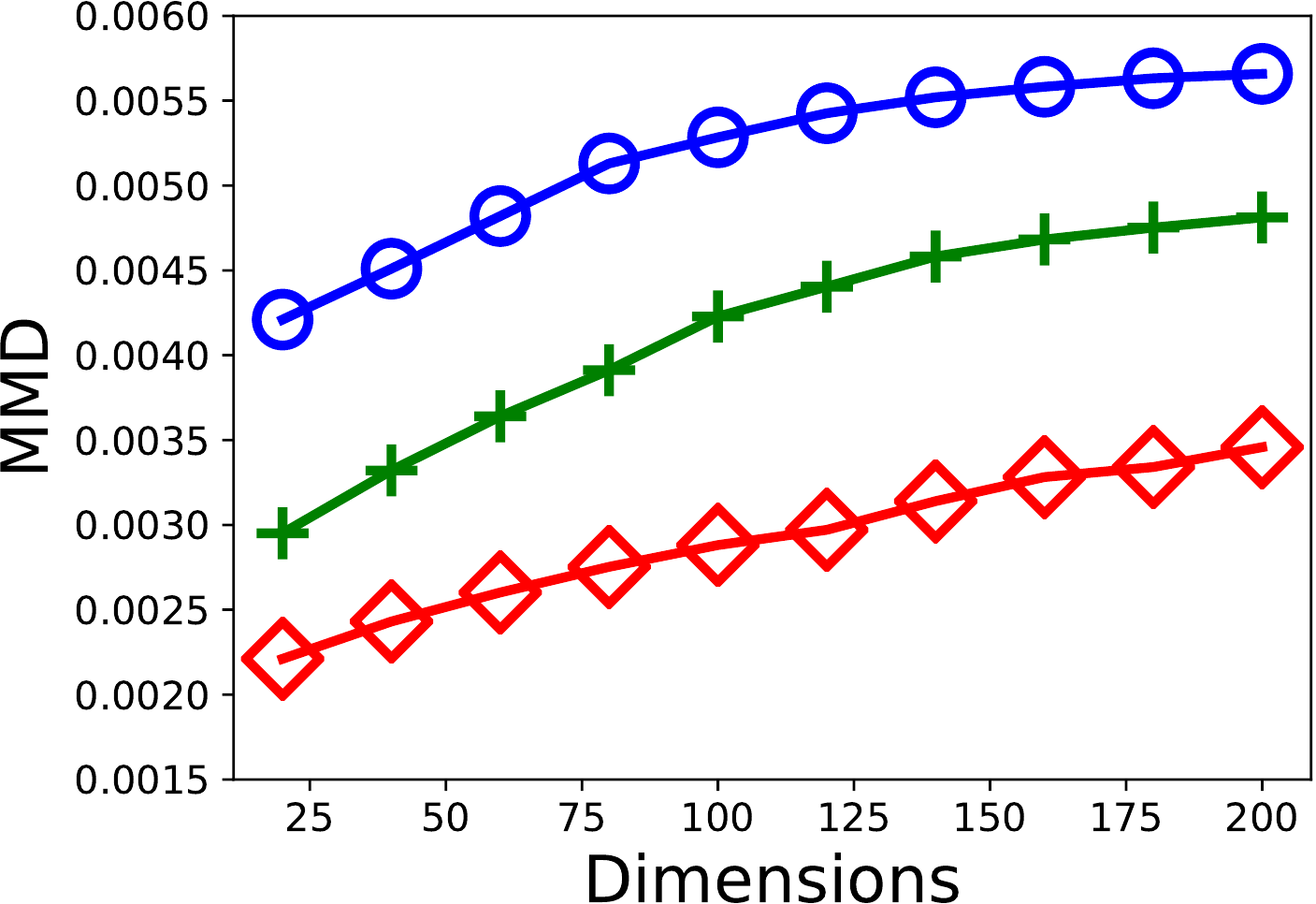} \\
{\small (a)  Fix dimension} &
{\small (b)  Fix sample size} 
\end{tabular}
\caption{\small Bernoulli RBM with number of visible units $M=25$. In (a), we fix the dimension of visible variables $d=100$ and vary the number of samples $\{\vz^j\}_{j=1}^n$. In (b), we fix the number of samples $n=100$ and vary the dimension of visible variables $d$. We calculate the MMD between the sample of different methods and the ground-truth sample. MSE is provided on Appendix.}
\label{fig:rbm}
\end{figure}

We compare our algorithm with Gibbs sampling and discontinuous HMC(DMHC, \cite{nishimura2017discontinuous}). In Fig.~\ref{fig:rbm}, $W$ is drawn from $N(0, 0.05)$, both $b$ and $c$ are drawn from $N(0, 1).$
With $10^5$ iterations of Gibbs sampling, we draw 500 parallel chains to take the last sample of each chain to get 500 ground-truth samples. We run Gibbs, DHMC and GF-SVGD at 500 iterations for fair comparison. In Gibbs sampling, $p(\vz\mid \vh)$ and $p(\vh\mid\vz)$ are iteratively sampled. In DHMC, a coordinate-wise integrator with Laplace momentum is applied to update the discontinuous states. We calculate MMD \cite{gretton2012kernel} between the ground truth sample and the sample drawn by different methods. The kernel in MMD is the exponentiated Hamming kernel from \cite{yang2018goodness}, defined as,
$k(\vz, \vz')=\exp(-H(\vz, \vz')),$ where $H(\vz, \vz'):=\frac{1}{d}\sum_{i=1}^d \mathbb{I}_{\{z_i \neq z_i'\}}$ is normalized Hamming distance. We perform experiments by fixing $d=100$ and varying sample size in Fig.~\ref{fig:rbm}(a) and fixing $n=100$ and varying $d$. Fig.~\ref{fig:rbm}(a) indicates that the samples from our method match the ground truth samples better in terms of MMD. Fig.~\ref{fig:rbm}(b) shows that the performance of our method is least sensitive to the dimension of the model than that of Gibss and DHMC. Both  Fig.~\ref{fig:rbm}(a) and  Fig.~\ref{fig:rbm}(b) show that our algorithm converges fastest.     
\begin{figure*}[tbh]
\begin{center}
\begin{tabular}{cccc}
\includegraphics[width=0.23\textwidth]{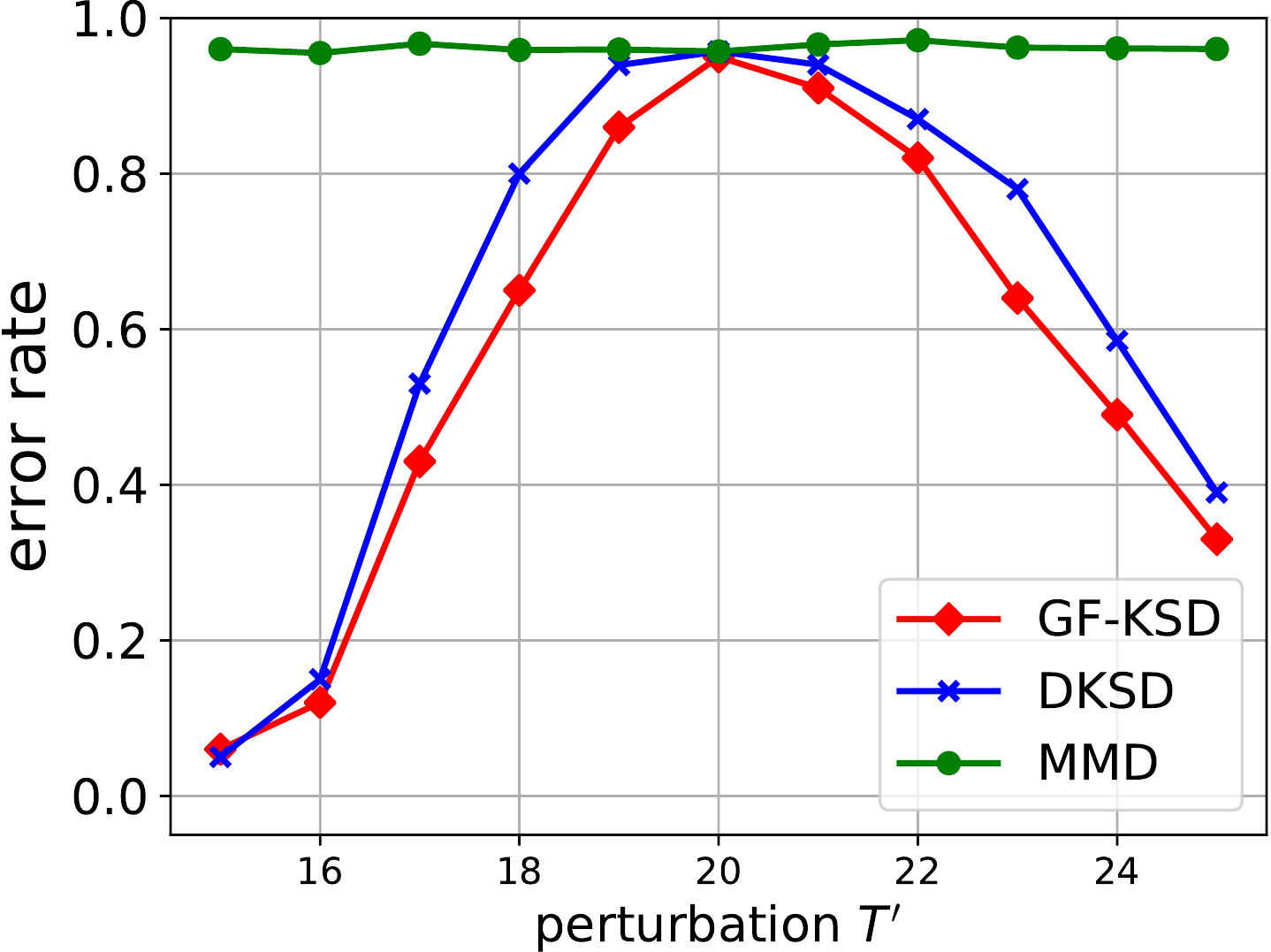} &
\hspace{-0.35cm}
\includegraphics[width=0.23\textwidth]{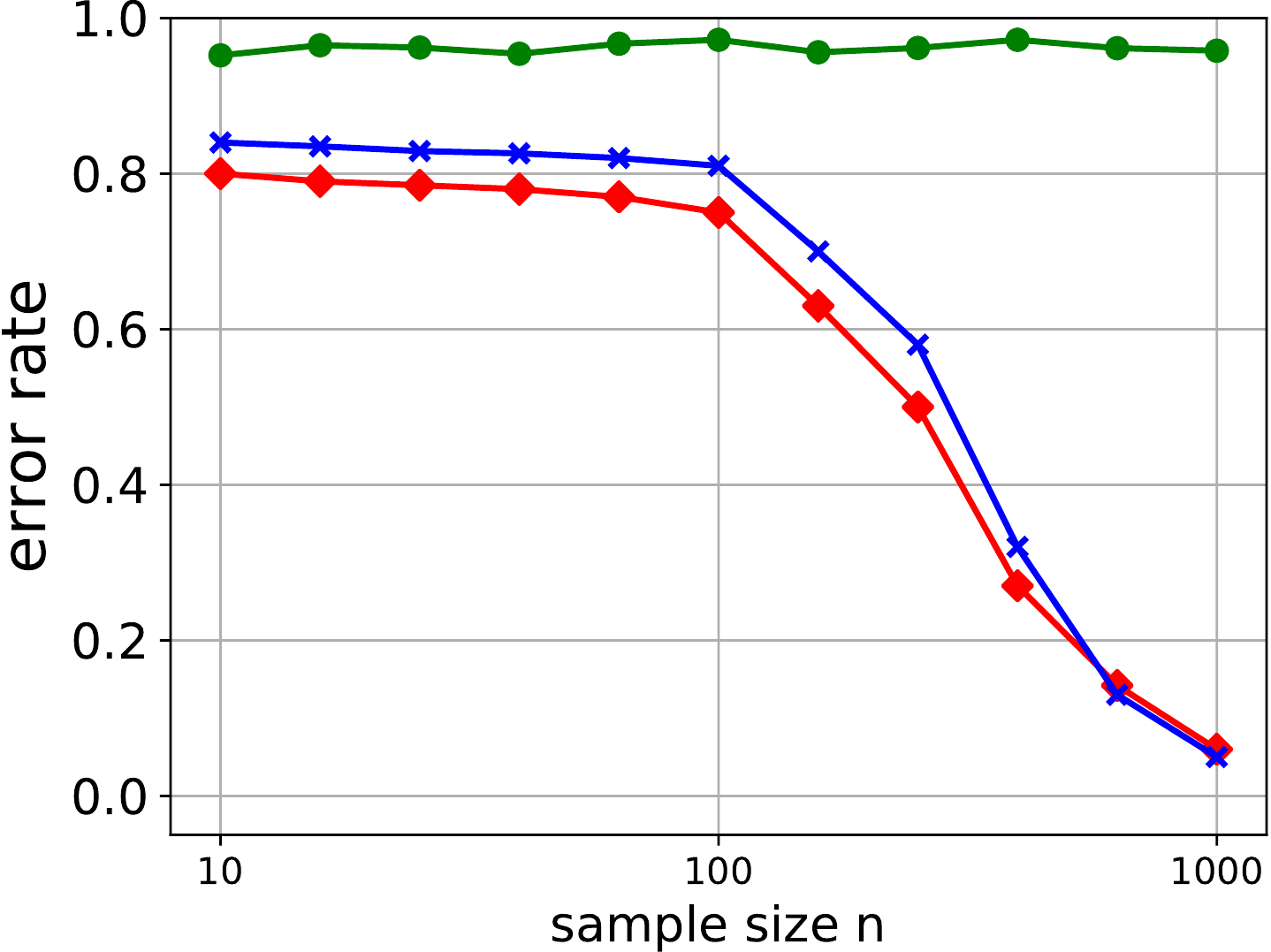} &
\hspace{-0.35cm}
\includegraphics[width=0.23\textwidth]{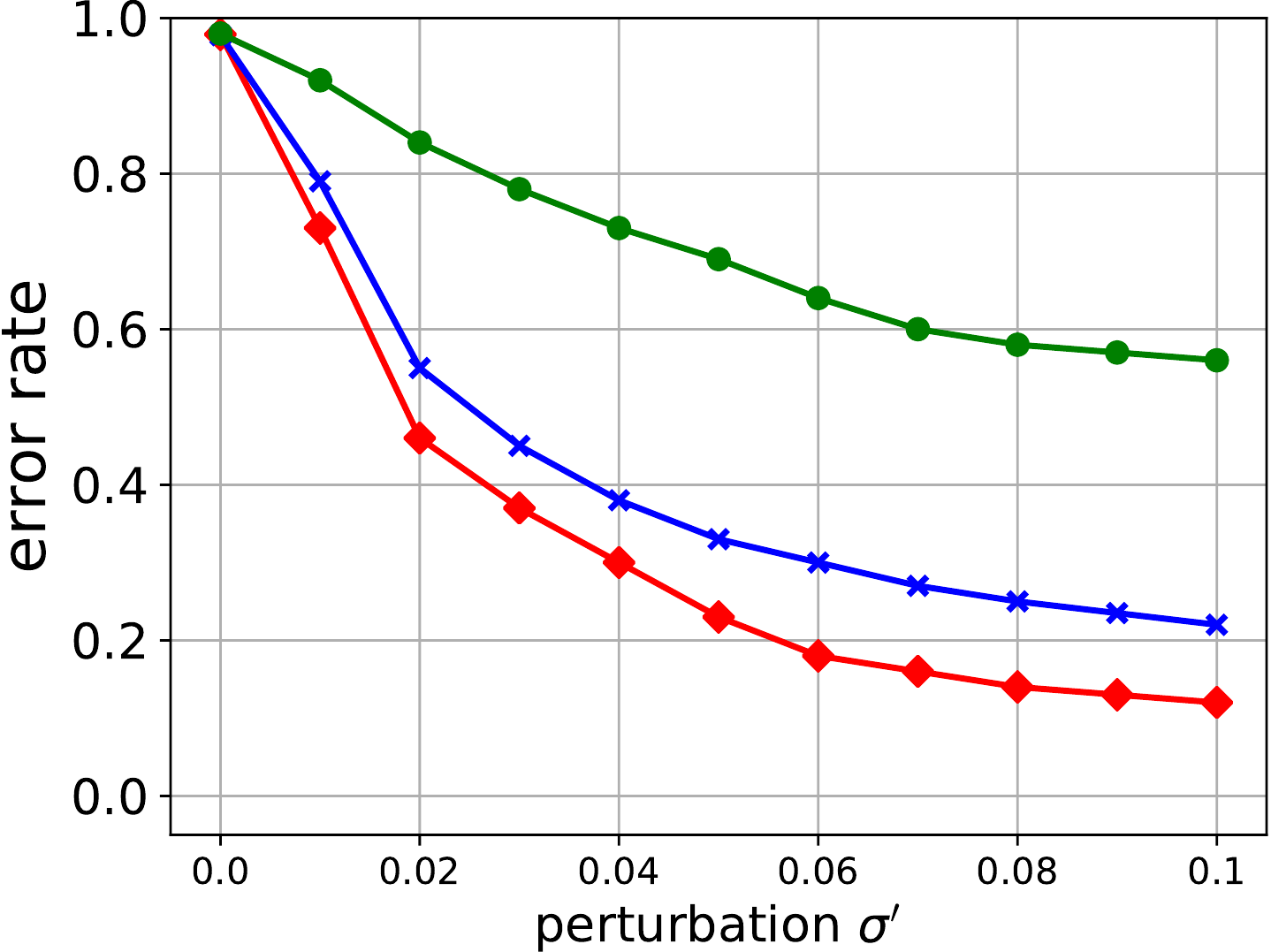} &
\hspace{-0.35cm}
\includegraphics[width=0.23\textwidth]{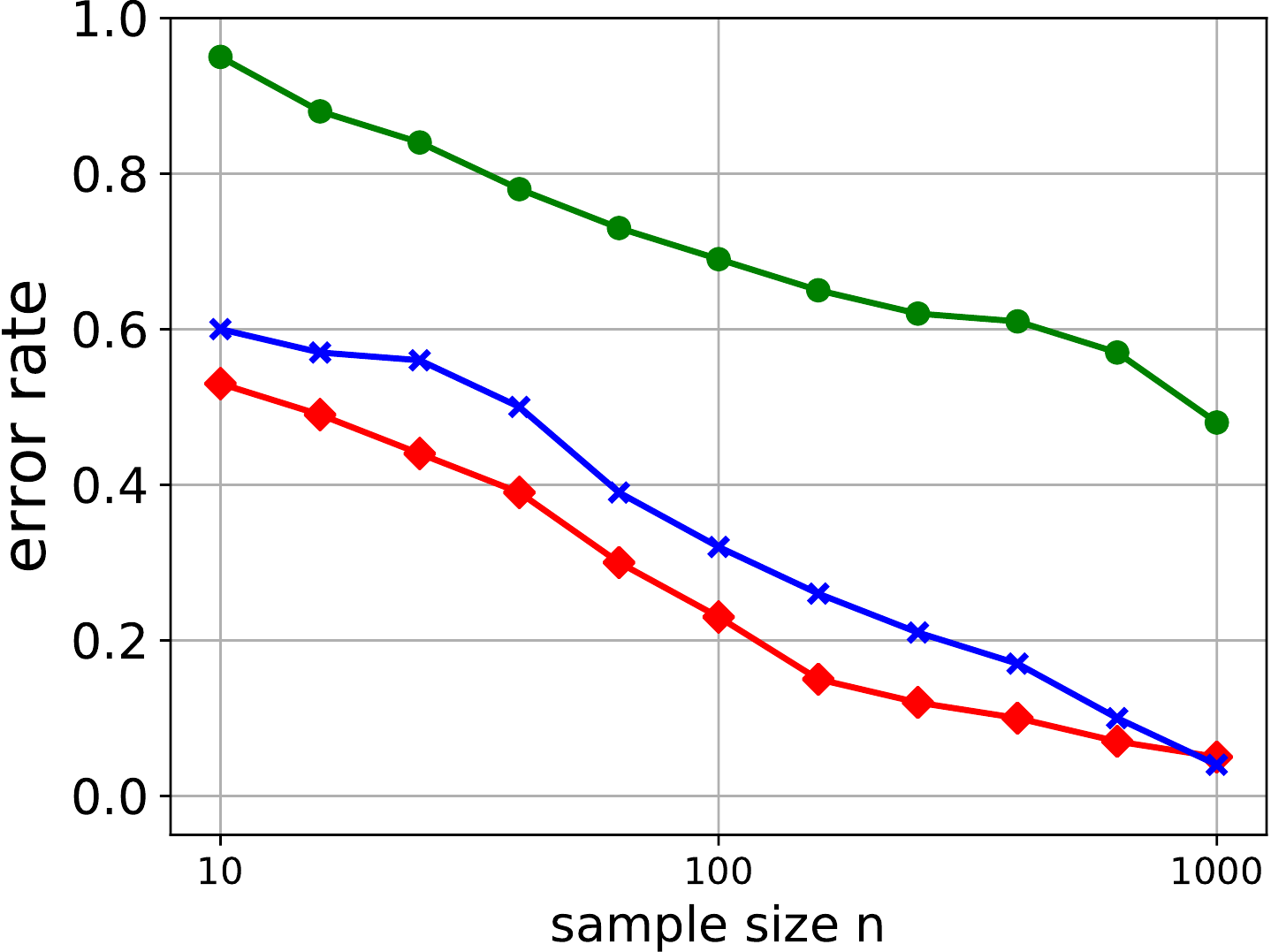}\\
{\small (a) $T=20$~(n=1000)} &
{\small (b) $T=20$ and $T'=15$}&
{\small (c) $\sigma=0$~(n=100)}&
{\small (d) $\sigma=0$ and $\sigma'=15$}
\end{tabular}
\caption{Goodness-of-fit test on Ising model (a, b) and Bernoulli RBM (c, d) with significant level $\alpha=0.05$. In (a, b), $p_*$ and $q_*$ has temperature $T$ and $T'$ respectively. In (c, d), $p_*$ has $W\sim \mathcal{N}(0, 1/M)$ and $q_*$ has $W+\epsilon$, where $\epsilon\sim \mathcal{N}(0, \sigma').$ $b$ and $c$ in $p_*$ and $q_*$ are the same. In (a, c) we vary the parameters of $q_*$. We fix the models and vary the sample size $n$ in (b, d). We test $H_0:q_*=p_*$ vs. $H_1:q_*\neq p_*.$ \label{fig:gof}}
\end{center}
\end{figure*}

\subsection{Learning Binarized Neural Network}
\begin{figure}[htb]
\centering
\begin{tabular}{cc}
\includegraphics[height=.2\textwidth]{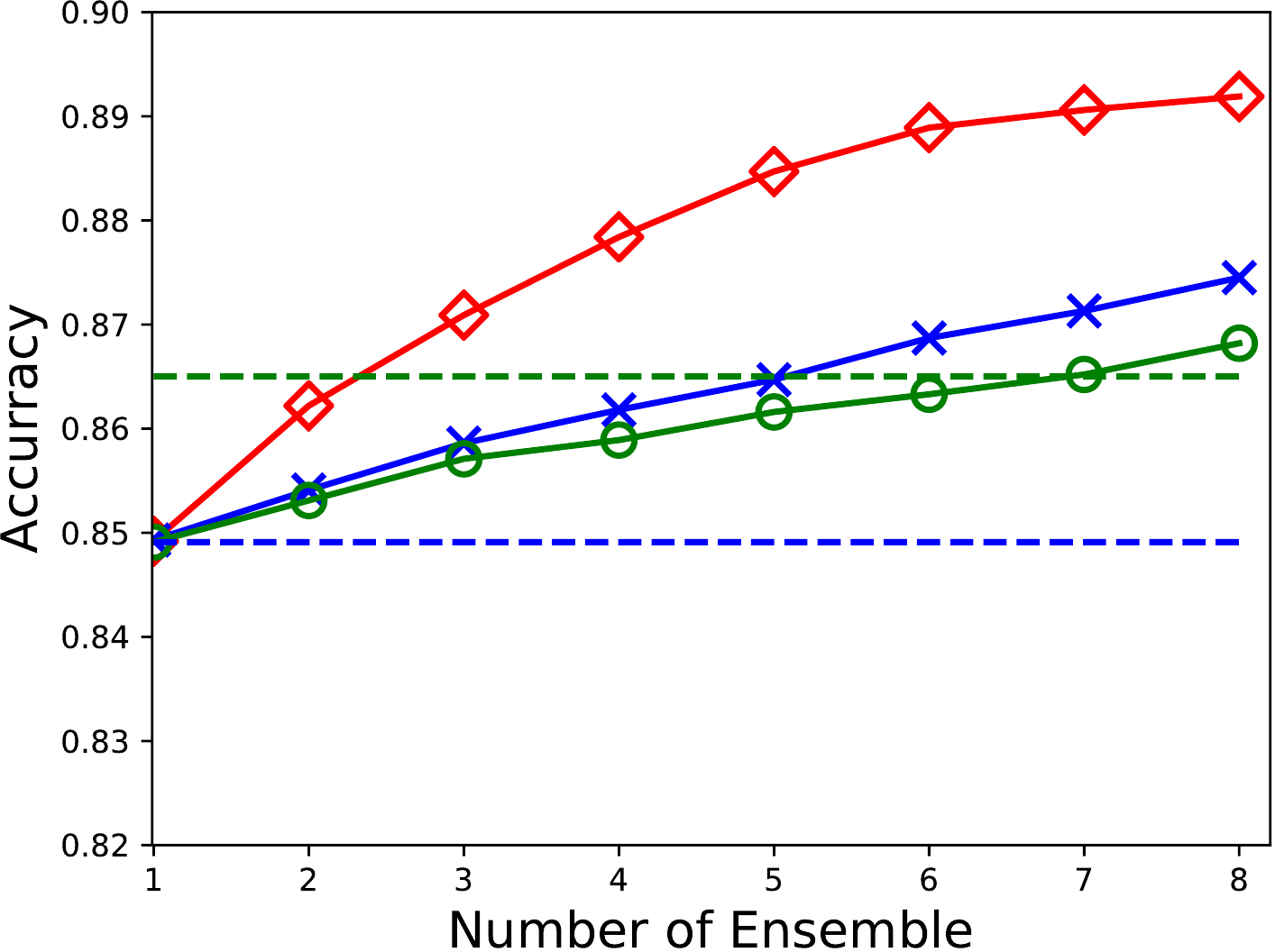}&
\raisebox{2em}{\includegraphics[height=0.1\textwidth]{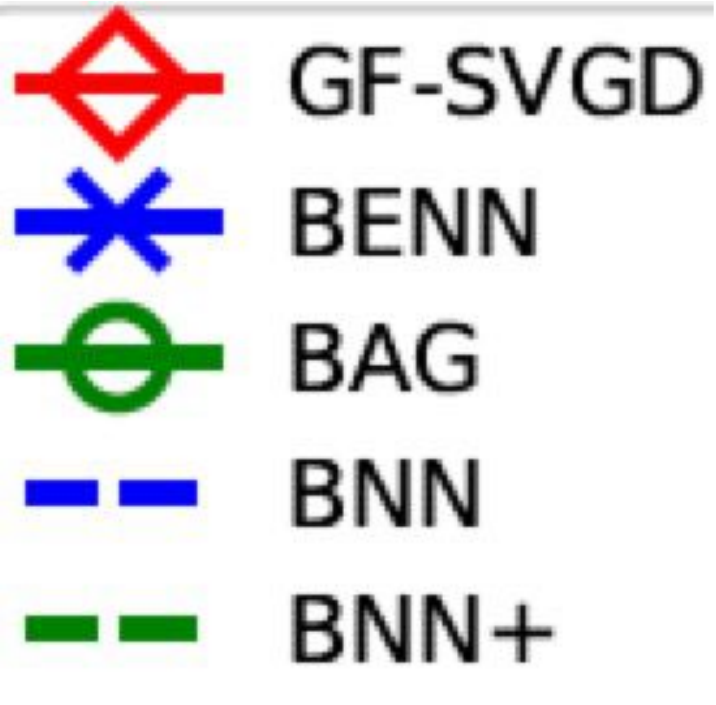}}
\end{tabular}
\caption{Comparison of different methods using AlexNet with binarized weights and activation on CIFAR10 dataset. We compare our GF-SVGD with BNN \cite{hubara2016binarized}, BNN+\cite{darabi2018bnn+} and BENN \cite{zhu2018binary}. "BAG" denote models are independently trained and linearly averaged the softmax output for prediction. Performance is based on the accuracy of different models w.r.t. ensemble size $n$ on test data.}
\label{fig:bnn}
\end{figure}
We slightly modify our algorithm to train binarized neural network (BNN), where both the weights and activation functions are binary $\pm 1$. BNN has been studied extensively because of its fast computation, energy efficiency and low memory cost \cite{rastegari2016xnor, hubara2016binarized, darabi2018bnn+, zhu2018binary}. The challenging problem in training BNN is that the gradients of the weights cannot be backpropagated through the binary activation functions because the gradients are zero almost everywhere. 

We train an ensemble of $n$ neural networks (NN) with the same architecture ($n\ge 2$). Let $\vw_i^b$ be the binary weight of model $i$, for $i=1,\cdots, n$, and $p_*(\vw_i^b;D)$ be the target probability model with softmax layer as last layer given the data $D$. Learning the target probability model is framed as drawing $n$ samples $\{\vw_i^b\}_{i=1}^n$ to approximate the posterior distribution $p_*(\vw^b; D)$. We train an ensemble of $n$ neural networks (NN) with the same architecture ($n\ge 2$). Let $\vw_i^b$ be the binary weight of model $i$, for $i=1,\cdots, n$, and $p_*(\vw_i^b;D)$ be the target probability model with softmax layer as last layer given the data $D$. Learning the target probability model is framed as drawing $n$ samples $\{\vw_i^b\}_{i=1}^n$ to approximate the posterior distribution $p_*(\vw^b; D)$. This involves sampling $\{\vw_i^b\}_{i=1}^n$ from discrete distributions $p_*(\vw^b; D)$, where our proposed sampling algorithm can be applied. Please refer to Appendix \ref{app:bnn} for the detail.

We test our ensemble algorithm by using binarized AlexNet \cite{krizhevsky2012imagenet} on CIFAR-10 dataset. We use the same setting for AlexNet as that in \cite{zhu2018binary}, which can be found in Appendix \ref{app:nndetail}. We compare our ensemble algorithm with typical ensemble method using bagging and AdaBoost (BENN, \cite{zhu2018binary}), BNN \cite{hubara2016binarized} and BNN+\cite{darabi2018bnn+}. Both BNN and BNN+ are trained on a single model with same network. From Fig.~\ref{fig:bnn}, we can see that all three ensemble methods (GF-SVGD, BAG and BENN) improve test accuracy over one single model (BNN and BNN+). To use the same setting for all methods, we don't use data augmentation or pre-training. Our ensemble method has the highest accuracy among all three ensemble methods. This is because our ensemble model are sufficiently interactive during training and our ensemble models $\{\vw_i\}$ in principle are approximating the posterior distribution $p(\vw;D).$

\subsection{Experiments on Goodness-of-fit Testing}
We perform goodness-of-fit tests on Ising model and Bernoulli RBM in Fig. \ref{fig:gof}, which shows type-II error rate (False negative error). The data $\{\vz_i\}_{i=1}^n$ is transformed to its corresponding continuous-valued samples $\{\vy_i\}_{i=1}^n,$ $y_i^j\in [0,\frac12),$ if $z_i^j=-1$; $y_i^j\in [\frac12, 1),$ if $z_i^j=1.$ Let $F$ be the c.d.f. of Gaussian base $p_0.$ By the same variable transform induced from $F,$ we obtain data $\vx^i=F^{-1}(\vy^i)$ and the transformed $p_c(\vx).$ The surrogate $\rho$ is chosen as that in sampling. Fig.~\ref{fig:gof} shows that our GF-KSD performs much better than DKSD~\cite{yang2018goodness} and MMD~\cite{gretton2012kernel} when the sample size $n$ is relatively small and the difference between $q_*$ and $p_*$ is within some range.

%% file: tex/conclusion.tex
\vspace{-.3cm}
\section{CONCLUSION}
In this paper, we propose a simple yet general framework to perform approximate inference and goodness-of-fit test on discrete distributions. We demonstrate the effectiveness of our proposed algorithm on a number of discrete graphical models. Based on our sampling method, we propose a new promising approach for learning an ensemble model of binarized neural networks. Future research includes applying our ensemble method to train BNN with larger networks such as VGG net and  larger dataset such as ImageNet dataset and extending our method to learn deep generative models with discrete distributions.

%% file: tex/appendix.tex
\begin{appendices}
\section*{Appendix}
\section{Additional Experimental Result\label{app:exp}}
\begin{figure*}[ht]
\centering
\subfigure[0th iteration]{\label{fig:binaryiter0}
\includegraphics[width=0.176\linewidth]{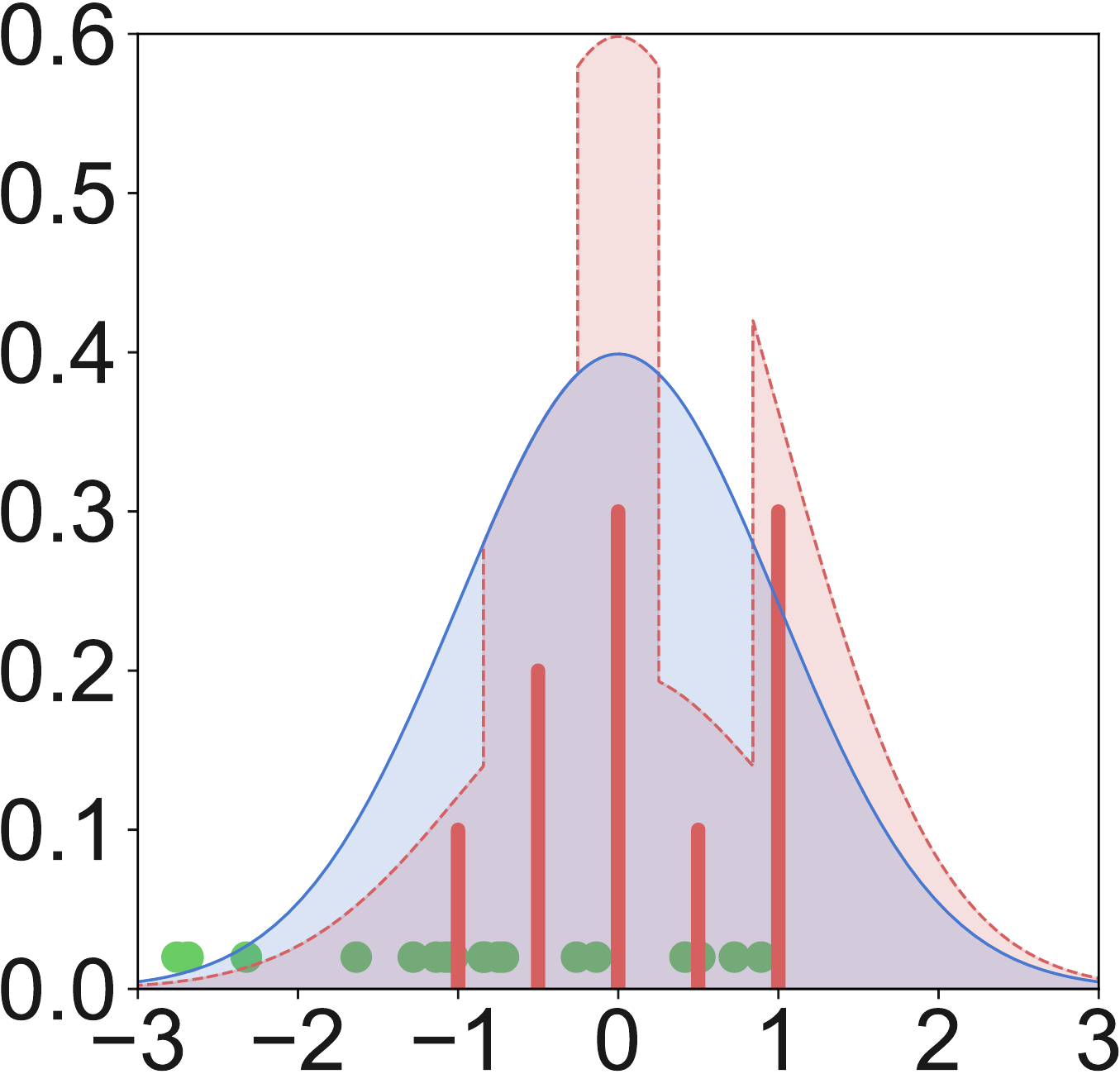}}
\subfigure[25th iteration]{\label{fig:binaryiter250}
\includegraphics[width=0.18\linewidth]{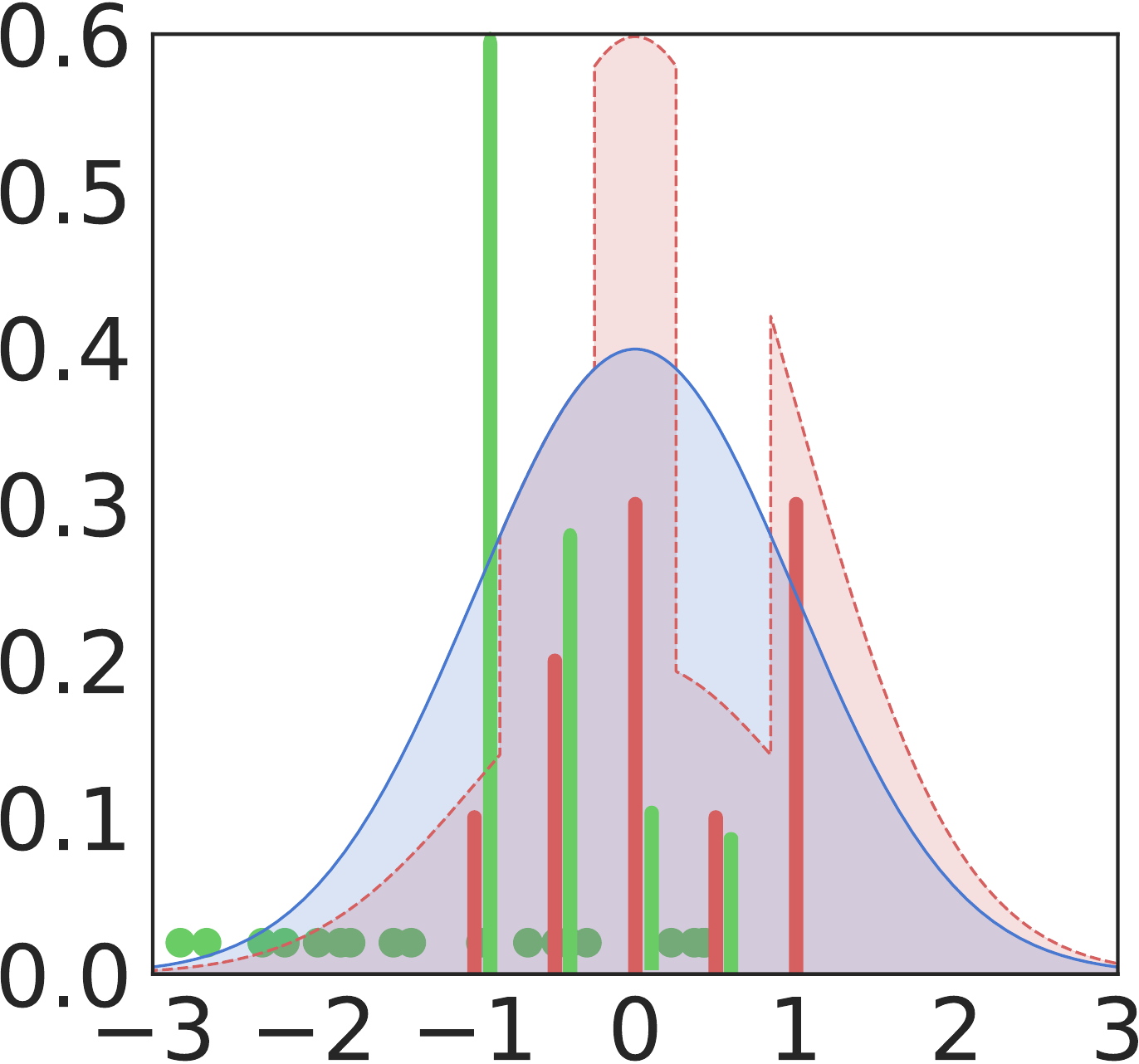}}
\subfigure[50th iteration]{\label{fig:binaryiter500}
\includegraphics[width=0.18\linewidth]{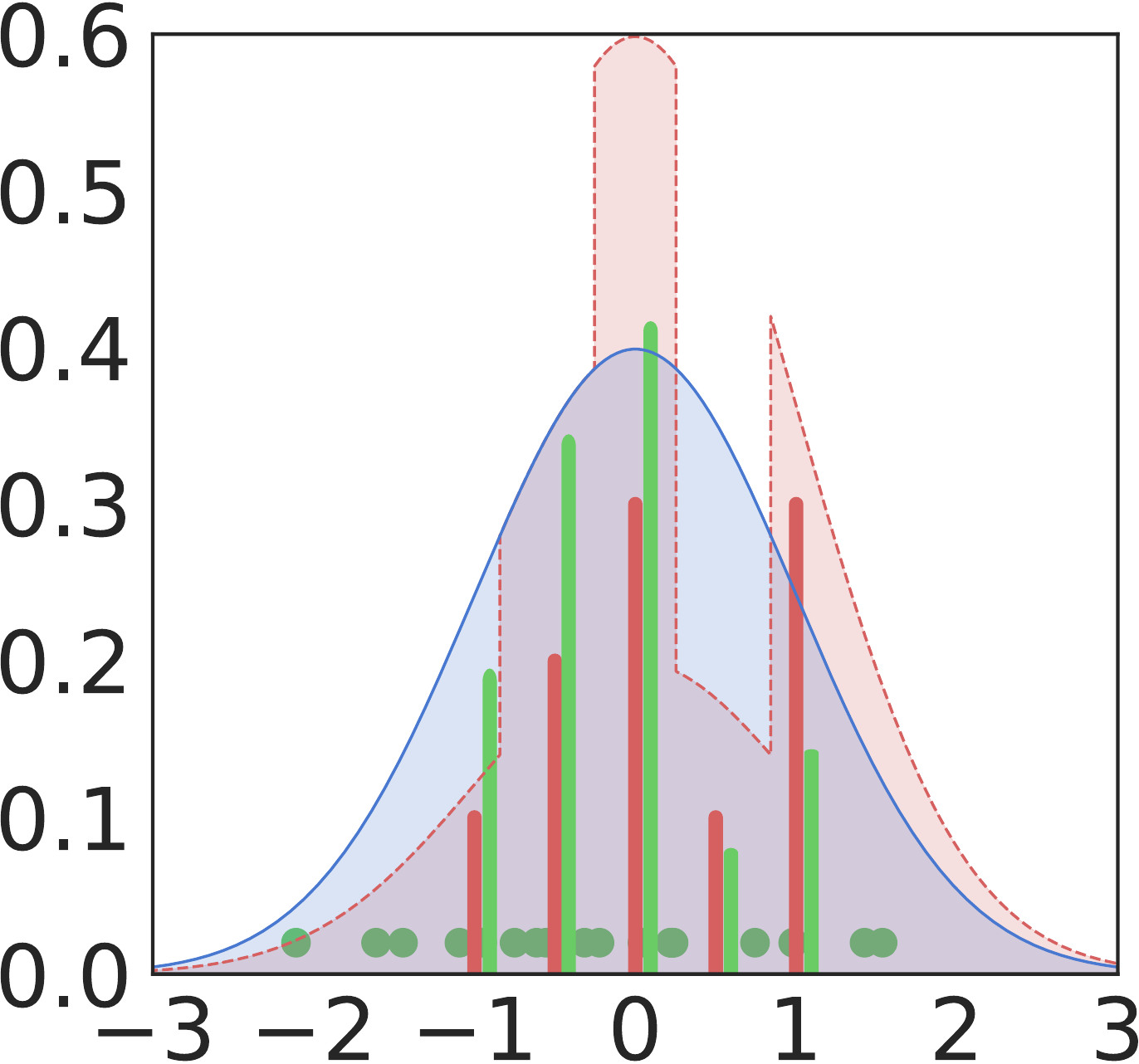}}
\subfigure[100th iteration]{\label{fig:binaryiter1000}
\includegraphics[width=0.18\linewidth]{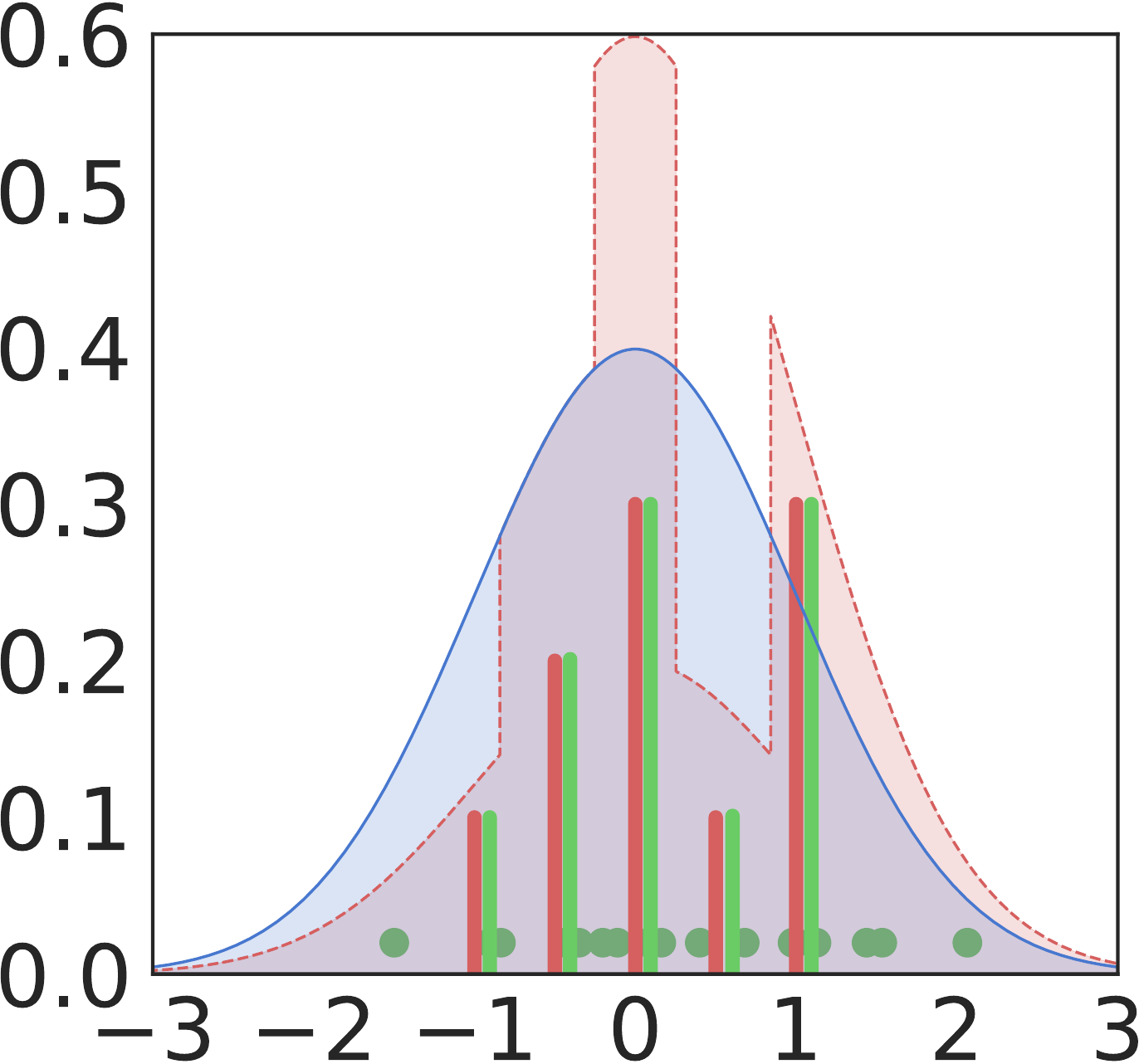}}
\subfigure{\label{fig:movement_legend}
\includegraphics[width=0.19\linewidth]{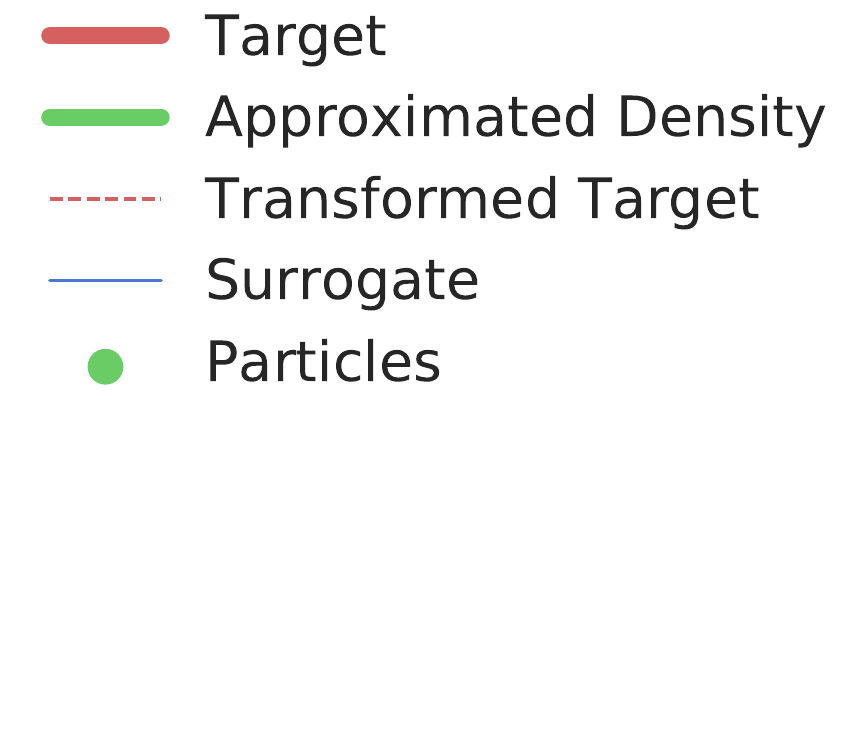}}
\caption{Evolution of real-valued particles $\{x_i\}_{i=1}^n$ (in green dots) by our discrete sampler in Alg.\ref{alg:alg1} on a one-dimensional categorical distribution. (a-d) shows particles $\{x^i\}$ at iteration 0, 10, 50 and 100 respectively. The categorical distribution is defined on states $z\in\{-1,-0.5, 0, 0.5, 1\}$ denoted by $a_1$, $a_2$, $a_3$, $a_4$, $a_5$, with probabilities $\{0.1, 0.2, 0.3, 0.1, 0.3\}$ denoted by $c_1$, $c_2$, $c_3$, $c_4$, $c_5$, respectively. $p_*(z=a_i)=c_i$. The base function is $p_0(x)$, shown in blue line. The transformed target to be sampled $p_c(x)\propto p_0(x)p_*(\Gamma(x))$, where $\Gamma(x)=a_i$ if $x\in[\eta_{i-1}, \eta_i)$ and $\eta_i$ is $i/5$-th quantile of standard Gaussian distribution. The surrogate distribution $\rho(x)$ is chosen as $p_0(x)$. We obtain discrete samples $\{z_i\}_{i=1}^n$ by $z_i=\Gamma(x_i)$. \label{fig:cat}}
\vspace{-0.1in}
\end{figure*}

\paragraph{Result on Categorical Distribution} We apply our algorithm to sample from one-dimensional categorical distribution $p_*(z)$ 
shown in red bars in Fig.~\ref{fig:cat}, 
defined on $\mathcal Z := \{-1,-0.5, 0, 0.5, 1\}$ with corresponding  probabilities $\{0.1, 0.2, 0.3,0.1, 0.3\}$.  The blue dash line is the surrogate distribution $\rho(x)=p_0(x)$, where the base function $p_0(x)$ is the p.d.f. of standard Gaussian distribution. The red dash line is the transformed piecewise continuous density $p_c(x)\propto p_0(x)p_*(\Gamma(x))$, where $\Gamma(x)=a_i$ if $x\in[\eta_{i-1}, \eta_i)$ and $\eta_i$ is $i/5$-th quantile of standard Gaussian distribution. We apply Algorithm~\ref{alg:alg1} to draw a set of samples $\{x_i\}_{i=1}^n$ (shown in green dots) to approximate the transformed target distribution. Then we can obtain a set of samples $\{z_i\}_{i=1}^n$ by $z_i=\Gamma(x_i))$, to get an approximation of the original categorical distribution.

As shown in Fig~\ref{fig:cat},
the empirical distribution
 of the discretized sample 
 $\{\vz_i\}_{i=1}^n$ (shown in green bars) 
aligns closely with the true distribution (the red bars) when the algorithm converges (e.g., at the 100-th iteration).  

\begin{figure}[h]
\centering
\begin{tabular}{cc}
\includegraphics[height=0.22\textwidth]{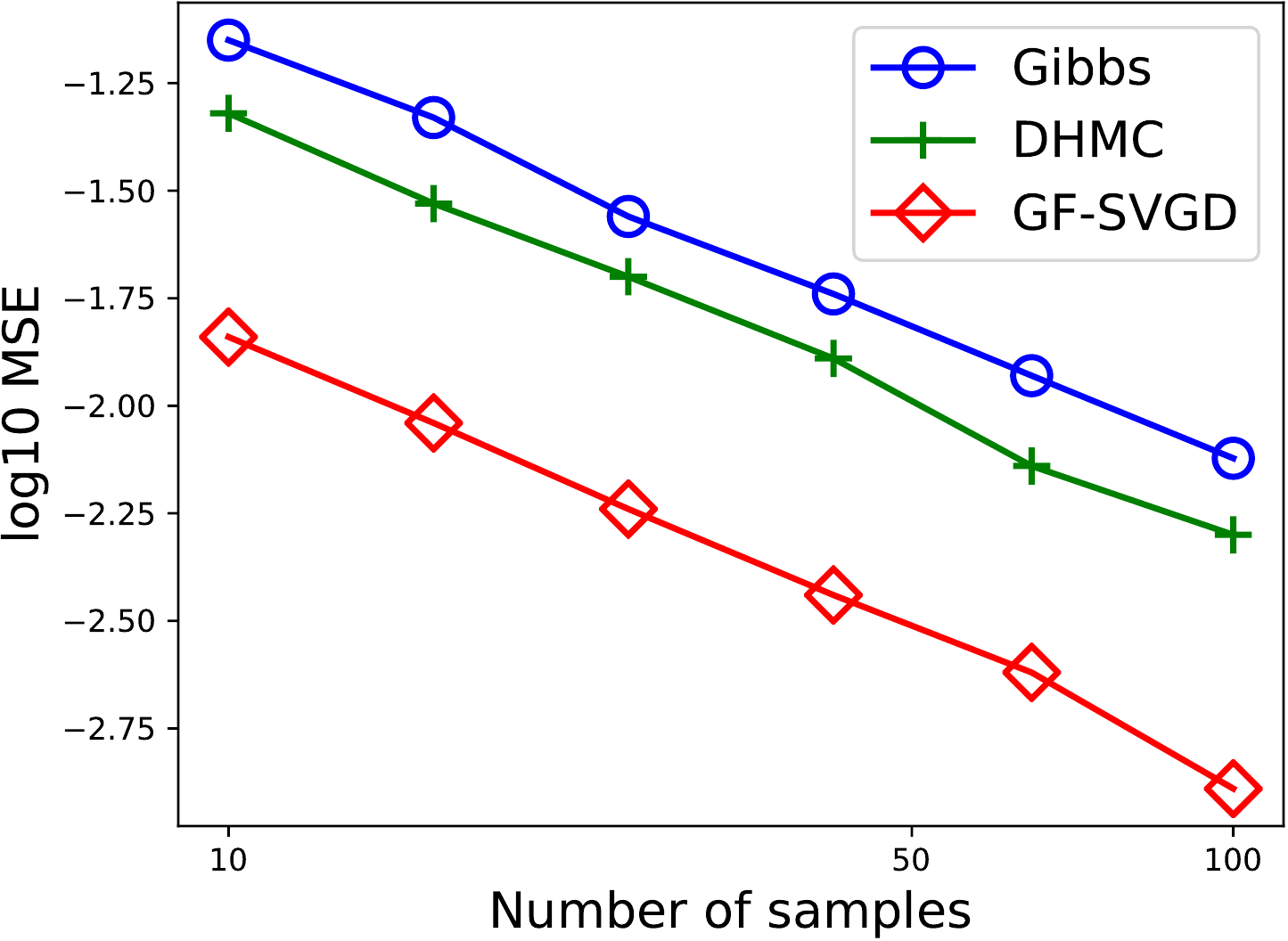} &
\includegraphics[height=0.22\textwidth]{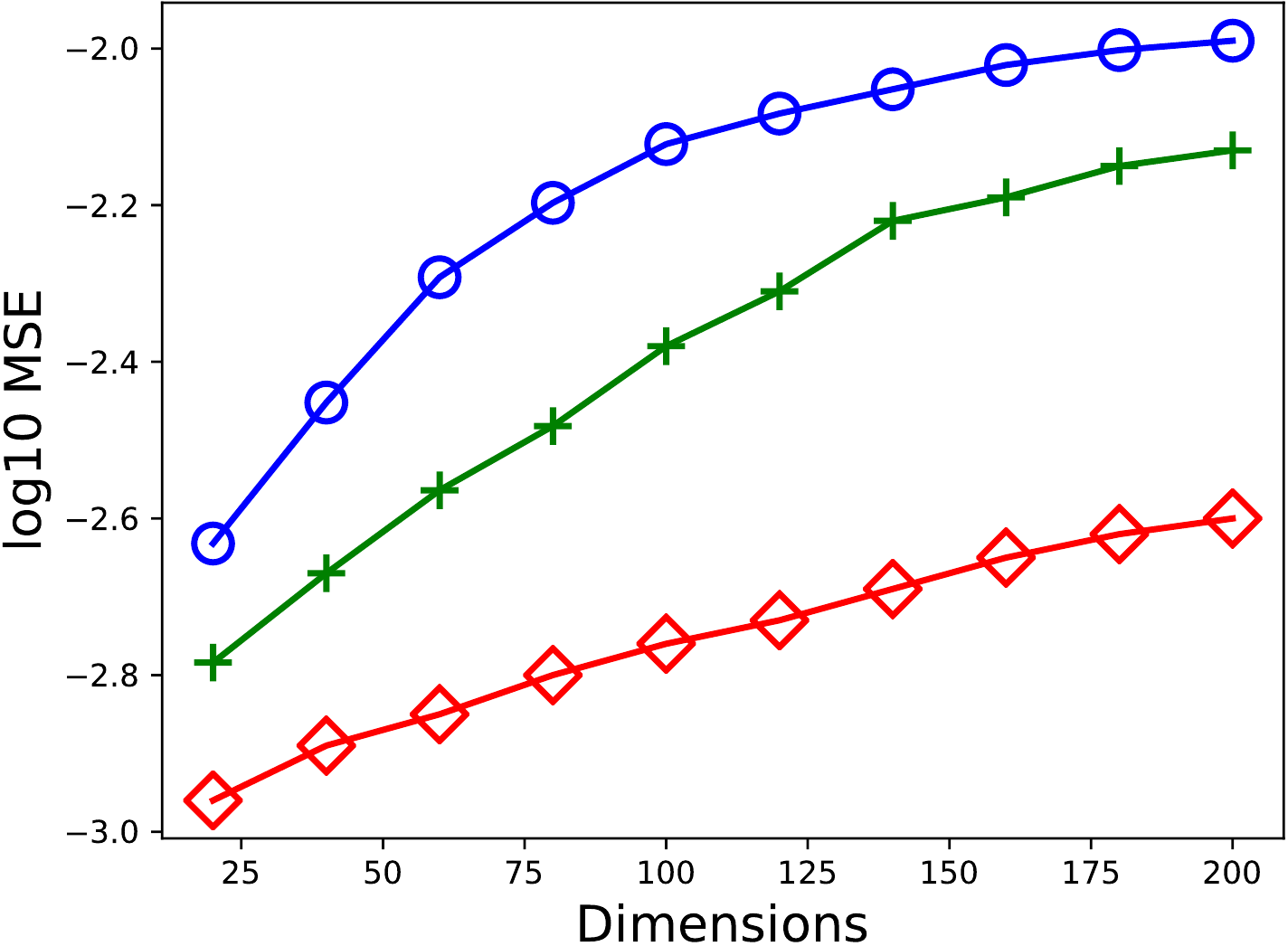} \\
{\small (a)  Fix dimension} &
{\small (b)  Fix sample size} 
\end{tabular}
\caption{\small Bernoulli RBM with number of visible units $M=25$. In (a), we fix the dimension of visible variables $d=100$ and vary the number of samples $\{\vz^j\}_{j=1}^n$. In (b), we fix the number of samples $n=100$ and vary the dimension of visible variables $d$. We calculate the MSE for estimating the mean $\E[z]$ (lower is better).\label{fig:rbm:mse}}

\end{figure}

\paragraph{Results on Bernoulli RBM}
The probability model is given in \eqref{def:rbm} and the score function is derived in Section 5.3~\cite{han2017stein}. We also evaluate the sample quality based on the mean square error (MSE) between the estimation and the ground truth value. From Fig.~\ref{fig:rbm:mse}(a), we can see that when fixing the dimension of the distribution $p_*(\vz)$, our sampling method has much lower MSE than Gibbs and DHMC. In Fig.~\ref{fig:rbm:mse}(b), as the dimension of the model increases, our sampling method has relatively better MSE than that of Gibbs and DHMC.  

\section{Training BNN Algorithm\label{app:bnn}}
In this section, we provide the procedure of our principled ensemble algorithm to train binarized neural network. We train an ensemble of $n$ neural networks (NN) with the same architecture ($n\ge 2$). Let $\vw_i^b$ be the binary weight of model $i$, for $i=1,\cdots, n$, and $p_*(\vw_i^b;D)$ be the target probability model with softmax layer as last layer given the data $D$. Learning the target probability model is framed as drawing $n$ samples $\{\vw_i^b\}_{i=1}^n$ to approximate the posterior distribution $p_*(\vw^b; D)$. We apply multi-dimensional transform $\vF$ to transform the original discrete-valued target to the target distribution of real-valued $\vw\in\R^d$. Let $p_0(w)$ be the base function, which is the product of the p.d.f. of the standard Gaussian distribution over the dimension $d.$ Based on the derivation in Section 3, the distribution of $\vw$ has the form $p_c(\vw;D)\propto p_*(\sign(\vw);D)p_0(\vw)$ with weight $\vw$ and the $\sign$ function is applied to each dimension of $\vw$. To backpropagate the gradient to the non-differentiable target, we construct a surrogate probability model $\rho(\vw;D)$ which approximates $\sign(\vw)$ in the transformed target by $\sigma(\vx)$ and relax the binary activation function $\{-1, 1\}$ by $\sigma$, where $\sigma$ is defined by \eqref{binary:approx}, denoted by $\wt{p}(\sigma(\vw);D)p_0(\vw)$. Here $\wt{p}(\sigma(\vw);D)$ is a differentiable approximation of $p_*(\sign(\vw);D).$  Then we apply GF-SVGD to update $\{\vw_i\}$ to approximate the transformed target distribution of $p_c(\vw;D)$ of $\vw$ as follows, 
$\vw_i \leftarrow \vw_i+\frac{\epsilon_{i}}{\Omega}\Delta \vw_i$, $\forall i=1,\cdots, n,$
\begin{equation}\label{bnn:update}
 \Delta \vw_i \!\! \leftarrow \!\! \! \sum_{j=1}^n \! \gamma_j [\nabla_{\vw}\log \rho(\vw_j;\!D_i)k(\vw_j\!,\!\vw_i)
             +\!\nabla_{\vw_j} k(\vw_j\!,\!\vw_i)]   
\end{equation}
where $D_i$ is batch data $i$ and $\mu_j =\rho(\vw_j; D_i)/p_c(\vw_j;D_i)$, $H(t) \overset{\mathrm{def}}{=}\sum_{j=1}^n \mathbb{I}(\mu_j\ge t)/n$, $\gamma_j= (H(\vw_j))^{-1}$ and $\Omega=\sum_{j=1}^n \gamma_j$. Note that we don't need to calculate the cumbersome term $p_0(\vw)$ as it can be canceled from the ratio between the surrogate distribution and the transformed distribution.  In practice, we find a more effective way to estimate this density ratio denoted by $\gamma_j$. Intuitively, this corresponds to assigning each particle a weight according to the rank of its density ratio in the population. Algorithm \ref{alg:GF-SVGDonBNN} on Appendix \ref{app:bnn} can be viewed as a new form of ensemble method for training NN models with discrete parameters. 

\begin {algorithm}[tbh]
\caption {GF-SVGD on training BNN}
\label{alg:GF-SVGDonBNN}
\begin {algorithmic}
\STATE {\bf Inputs}: training set $D$ and testing set $D_{\mathrm{test}}$
\STATE {\bf Outputs}: classification accuracy on testing set.
\STATE {\bf Initialize} full-precision models $\{\vw^i\}_{i=1}^n$ and its binary form $\{\vw^b_i\}_{i=1}^n$ where $\vw_i^b=\sign(\vw^i)$.
\WHILE{not converge}
\STATE -Sample $n$ batch data $\{D_i\}_{i=1}^n.$
\STATE -Calculate the true likelihood $p_c(\vw_i; D_i)\propto p_*(\sign(\vw_i);D_i)p_0(x)$ 
\STATE -Relax $\vw^i_b$ with $\sigma(\vw_i)$
            \STATE -Relax each sign activation function to the smooth function defined in \eqref{binary:approx} to get $\wt{p}$
            \STATE -Calculate the surrogate likelihood $\rho(\vw^i;D_i)\propto\wt{p}(\sigma(\vw_i);D_{i})p_0(\vx)$
            \STATE -$\vw_i \leftarrow \vw_i+\Delta \vw_i$, $\forall i=1,\cdots, n,$ where $\Delta \vw_i$ is defined in \eqref{bnn:update}.
            \STATE -Clip $\{\vw_i\}$ to interval $(-1, 1)$ for stability.
\ENDWHILE            
\STATE  -Calculate the probability output by softmax layer $p(\vw_i^b;D_{\mathrm{test}})$
\STATE -Calculate the average probability $f(\vw_b;D_{\mathrm{test}})\leftarrow \sum_{i=1}^{n} p(\vw_i^b;D_{\mathrm{test}})$
\STATE {\bf Output} test accuracy from $f(\vw_b;D_{\mathrm{test}}).$
\end {algorithmic}
\end {algorithm}

\section{Transform Discrete Samples to Continuous Samples for Goodness-of-fit Test\label{app:gof:transf}} Let $F$ be the c.d.f. of Gaussian base density $p_0.$ Let us first illustrate how to transform one-dimensional samples $\{z_i\}_{i=1}^n$ to continuous samples.
\begin{enumerate}
  \vspace{-.3cm}
    \item Given discrete data $\{z_i\}_{i=1}^n.$ 
    Let $\{a_j\}_{j=1}^K$ are possible discrete states. Assume $K$ is large so that for any $z_i,$ we have $z_i=a_j$ for one $j.$
    \vspace{-.2cm}
    \item For any $ z_i$ such as $z_i=a_j$, randomly sample $y_i\in [\frac{j-1}{K}, \frac{j}{K}).$ We obtain data $\{y_i\}_{i=1}^n.$ 
    \vspace{-.2cm}
    \item Apply $x=F^{-1}(y),$ we obtain data $\{x_i\}_{i=1}^n.$
    \vspace{-.3cm}
\end{enumerate}
For $\vx=(x^1,\cdots, x^d),$ let $F(\vx)=(F_1(x^1),\cdots, F_d(x^d)$, where $F_i$ is the c.d.f. of Gaussian density $p_{0,i}(x^i).$ We apply the above one-dimensional transform to each dimension of $\{\vz_i\}_{i=1}^n,$ $\vz_i=(z_i^1, \cdots, z_i^d).$ We can easily obtain the continuous data $\{\vx_i\}_{i=1}^n.$

\section{Proofs}
In the following, we prove proposition 4.

{\bf Proposition 4} Assume $\Gamma$ is an even partition of $p_0(\vx)$, and $p_c(\vx) = K p_0(\vx) p_*(\Gamma(\vx))$, where $K$ severs as a normalization constant, then $(p_c, ~ \Gamma)$ is a continuous parameterisation of $p_*$. 
\begin{proof}
We just need to verify that \eqref{pstar} holds. 
\begin{align*}
    & \int p_c(\vx) \ind[\va_i = \Gamma(
    \vx)] d\vx  \\
    & = K \int p_0(\vx) p_*(\Gamma(\vx)) \ind[a_i = \Gamma(\vx)] d\vx   \\
    & = K \int p_0(\vx) p_*(\va_i) \ind[\va_i = \Gamma(\vx)] d\vx \\
    & = K p_*(\va_i) \int p_0(\vx)  \ind[\va_i = \Gamma(\vx)] d\vx \\
    & = p_*(\va_i), 
\end{align*} 
where the last step follows \eqref{even}. 
\end{proof}

\section{Detail of Experiments and Network Architecture\label{app:nndetail}}
In all experiments, we use RBF kernel $k(\vx, \vx')=\exp(-\|\vx-\vx'\|^2/h)$ for the updates of our proposed algorithms; the bandwidth $h$ is taken to be $h {=} \mathrm{med^2}/(2\log(n+1))$ where $\mathrm{med}$ is the median of the current $n$ particles. Adam optimizer \cite{kingma2014adam} is applied to our proposed algorithms for accelerating convergence. $\epsilon=0.0001$ works for all the experiments. 

We use the same AlexNet as \cite{zhu2018binary}, which is illustrated in the following.
{\small 
\begin{table}[h]
    \centering
    \begin{tabular}{|c|c|c|} \hline
    Layer & Type & Parameters \\\hline
     1    & Conv & Depth: 96, K: $11\times 11$, S: 4, P:0 \\
     2  &  Relu &  - \\
     3 & MaxPool & K: $3\times 3$, S: 2 \\
     4 & BatchNorm & -  \\
     5 & Conv & Depth: 256, K: $5\times 5$, S: 1, P:1 \\
    6  &  Relu &  - \\
      7 & MaxPool & K: $3\times 3$, S: 2 \\
     8 & BatchNorm & -  \\
     9 & Conv & Depth: 384, K: $3\times 3$, S: 1, P:1 \\
     10  &  Relu &  - \\
     11 & Conv & Depth: 384, K: $3\times 3$, S: 1, P:1 \\
    12  &  Relu &  - \\
    13 & Conv & Depth: 256, K: $3\times 3$, S: 1, P:1 \\
    14  &  Relu &  - \\
    15 & MaxPool & K: $3\times 3$, S: 2 \\
    16  &Dropout & $p=0.5$\\
    17  & FC & Width=4096\\
     18  &  Relu &  - \\
    19  &Dropout & $p=0.5$\\
    20  & FC & Width=4096\\
    21  &  Relu &  - \\
    22 & FC & Width=10\\\hline
    \end{tabular}
    \caption{Architecture of AlexNet. "K" denotes kernel size; "S" denotes stride; "P" denotes padding.}
    \label{tab:alexnet}
\end{table}
}

\end{appendices}

%% file: DiscreteSVGDUAI.bbl
\begin{thebibliography}{10}\itemsep=-1pt

\bibitem{afshar2015reflection}
H.~M. Afshar and J.~Domke.
\newblock Reflection, refraction, and hamiltonian monte carlo.
\newblock In {\em NIPS}, 2015.

\bibitem{ahn2016synthesis}
S.-S. Ahn, M.~Chertkov, and J.~Shin.
\newblock Synthesis of mcmc and belief propagation.
\newblock In {\em Advances in Neural Information Processing Systems}, 2016.

\bibitem{chwialkowski2016kernel}
K.~Chwialkowski, H.~Strathmann, and A.~Gretton.
\newblock A kernel test of goodness of fit.
\newblock In {\em ICML}, 2016.

\bibitem{darabi2018bnn+}
S.~Darabi, M.~Belbahri, M.~Courbariaux, and V.~P. Nia.
\newblock Bnn+: Improved binary network training.
\newblock {\em arXiv:1812.11800}, 2018.

\bibitem{daskalakis2019testing}
C.~Daskalakis, N.~Dikkala, and G.~Kamath.
\newblock Testing ising models.
\newblock {\em IEEE Transactions on Information Theory}, 2019.

\bibitem{dechter1998bucket}
R.~Dechter.
\newblock Bucket elimination: A unifying framework for probabilistic inference.
\newblock In {\em Learning in graphical models}. Springer, 1998.

\bibitem{dinh2017probabilistic}
V.~Dinh, A.~Bilge, C.~Zhang, and F.~A. Matsen~IV.
\newblock Probabilistic path hamiltonian monte carlo.
\newblock In {\em ICML}, 2017.

\bibitem{gretton2012kernel}
A.~Gretton, K.~M. Borgwardt, M.~J. Rasch, B.~Sch{\"o}lkopf, and A.~Smola.
\newblock A kernel two-sample test.
\newblock {\em Journal of Machine Learning Research}, 13(Mar), 2012.

\bibitem{han2017stein}
J.~Han and Q.~Liu.
\newblock Stein variational adaptive importance sampling.
\newblock {\em arXiv preprint arXiv:1704.05201}, 2017.

\bibitem{han2018stein}
J.~Han and Q.~Liu.
\newblock Stein variational gradient descent without gradient.
\newblock {\em arXiv preprint arXiv:1806.02775}, 2018.

\bibitem{hinton2002training}
G.~E. Hinton.
\newblock Training products of experts by minimizing contrastive divergence.
\newblock {\em Neural computation}, 14(8):1771--1800, 2002.

\bibitem{hubara2016binarized}
I.~Hubara, M.~Courbariaux, D.~Soudry, R.~El-Yaniv, and Y.~Bengio.
\newblock Binarized neural networks.
\newblock In {\em NIPS}, 2016.

\bibitem{ising1924beitrag}
E.~Ising.
\newblock {\em Beitrag zur theorie des ferro-und paramagnetismus}.
\newblock PhD thesis, Hamburg, 1924.

\bibitem{kingma2014adam}
D.~P. Kingma and J.~Ba.
\newblock Adam: A method for stochastic optimization.
\newblock {\em arXiv preprint arXiv:1412.6980}, 2014.

\bibitem{krizhevsky2012imagenet}
A.~Krizhevsky, I.~Sutskever, and G.~E. Hinton.
\newblock Imagenet classification with deep convolutional neural networks.
\newblock In {\em NIPS}, 2012.

\bibitem{liu2017stein}
Q.~Liu.
\newblock Stein variational gradient descent as gradient flow.
\newblock In {\em Advances in neural information processing systems}, 2017.

\bibitem{liu2015probabilistic}
Q.~Liu, J.~W. Fisher~III, and A.~T. Ihler.
\newblock Probabilistic variational bounds for graphical models.
\newblock In {\em Advances in Neural Information Processing Systems}, 2015.

\bibitem{liu2016kernelized}
Q.~Liu, J.~Lee, and M.~Jordan.
\newblock A kernelized stein discrepancy for goodness-of-fit tests.
\newblock In {\em International Conference on Machine Learning}, 2016.

\bibitem{liu2016stein}
Q.~Liu and D.~Wang.
\newblock Stein variational gradient descent: A general purpose bayesian
  inference algorithm.
\newblock In {\em NIPS}, pages 2378--2386, 2016.

\bibitem{lou2017dynamic}
Q.~Lou, R.~Dechter, and A.~T. Ihler.
\newblock Dynamic importance sampling for anytime bounds of the partition
  function.
\newblock In {\em Advances in Neural Information Processing Systems}, 2017.

\bibitem{martin2017exact}
A.~Mart{\'\i}n~del Campo, S.~Cepeda, and C.~Uhler.
\newblock Exact goodness-of-fit testing for the ising model.
\newblock {\em Scandinavian Journal of Statistics}, 2017.

\bibitem{nishimura2017discontinuous}
A.~Nishimura, D.~Dunson, and J.~Lu.
\newblock Discontinuous hamiltonian monte carlo for discrete parameters and
  discontinuous likelihoods.
\newblock {\em arXiv:1705.08510}, 2019.

\bibitem{pakman2013auxiliary}
A.~Pakman and L.~Paninski.
\newblock Auxiliary-variable exact hamiltonian monte carlo samplers for binary
  distributions.
\newblock In {\em NIPS}, pages 2490--2498, 2013.

\bibitem{rastegari2016xnor}
M.~Rastegari, V.~Ordonez, J.~Redmon, and A.~Farhadi.
\newblock Xnor-net: Imagenet classification using binary convolutional neural
  networks.
\newblock In {\em European Conference on Computer Vision}. Springer, 2016.

\bibitem{valiant2016instance}
G.~Valiant and P.~Valiant.
\newblock Instance optimal learning of discrete distributions.
\newblock In {\em Proceedings of the forty-eighth annual ACM symposium on
  Theory of Computing}. ACM, 2016.

\bibitem{wainwright2008graphical}
M.~J. Wainwright, M.~I. Jordan, et~al.
\newblock Graphical models, exponential families, and variational inference.
\newblock {\em Foundations and Trends{\textregistered} in Machine Learning},
  2008.

\bibitem{yang2018goodness}
J.~Yang, Q.~Liu, V.~Rao, and J.~Neville.
\newblock Goodness-of-fit testing for discrete distributions via stein
  discrepancy.
\newblock In {\em ICML}, 2018.

\bibitem{zhang2012continuous}
Y.~Zhang, Z.~Ghahramani, A.~J. Storkey, and C.~A. Sutton.
\newblock Continuous relaxations for discrete hamiltonian monte carlo.
\newblock In {\em NIPS}, 2012.

\bibitem{zhu2018binary}
S.~Zhu, X.~Dong, and H.~Su.
\newblock Binary ensemble neural network: More bits per network or more
  networks per bit?
\newblock {\em arXiv:1806.07550}, 2018.

\end{thebibliography}
